\documentclass[sigconf]{acmart}
\usepackage{subfigure}
\usepackage{multicol}
\usepackage{multirow}
\usepackage{tabularx}
\usepackage{diagbox}
\newtheorem{definition}{Definition}
\usepackage{hyperref}
\AtBeginDocument{%
  }

\copyrightyear{2025}
\acmYear{2025}
\setcopyright{acmlicensed}\acmConference[KDD '25]{Proceedings of the 31st ACM SIGKDD Conference on Knowledge Discovery and Data Mining V.2}{August 3--7, 2025}{Toronto, ON, Canada}
\acmBooktitle{Proceedings of the 31st ACM SIGKDD Conference on Knowledge Discovery and Data Mining V.2 (KDD '25), August 3--7, 2025, Toronto, ON, Canada}
\acmDOI{10.1145/3711896.3737157}
\acmISBN{979-8-4007-1454-2/2025/08}

\begin{document}

\title[TimeCapsule: Long-Term Time Series Forecasting with Compressed Predictive Representations]{TimeCapsule: Solving the Jigsaw Puzzle of Long-Term Time Series Forecasting with Compressed Predictive Representations}

\author{Yihang Lu}
\affiliation{%
  \institution{Hefei Institutes of Physical Science}
  \city{Hefei}
  \country{China}
}
\affiliation{%
  \institution{University of Science and Technology of China}
  \city{Hefei}
  \country{China}
}
\email{lyhsa22@mail.ustc.edu.cn}

\author{Yangyang Xu}
\affiliation{%
  \institution{Hefei Institutes of Physical Science}
  \city{Hefei}
  \country{China}
}
\affiliation{%
  \institution{University of Science and Technology of China}
  \city{Hefei}
  \country{China}
}
\email{xyy18356098009@mail.ustc.edu.cn}

\author{Qitao Qing}
\affiliation{%
  \institution{University of Science and Technology of China}
  \city{Hefei}
  \country{China}
}
\email{qqt@mail.ustc.edu.cn}
\author{Xianwei Meng}
\authornote{Corresponding Author.}
\affiliation{%
  \institution{Hefei Institutes of Physical Science}
  \city{Hefei}
  \country{China}
}
\email{mengxw@iim.ac.cn}

\begin{abstract}
  Recent deep learning models for Long-term Time Series Forecasting (LTSF) often emphasize complex, handcrafted designs, while simpler architectures like linear models or MLPs have often outperformed these intricate solutions. In this paper, we revisit and organize the core ideas behind several key techniques, such as redundancy reduction and multi-scale modeling, which are frequently employed in advanced LTSF models. Our goal is to streamline these ideas for more efficient deep learning utilization. To this end, we introduce TimeCapsule, a model built around the principle of high-dimensional information compression that unifies these techniques in a generalized yet simplified framework. Specifically, we model time series as a 3D tensor, incorporating temporal, variate, and level dimensions, and leverage mode production to capture multi-mode dependencies while achieving dimensionality compression. We propose an internal forecast within the compressed representation domain, supported by the Joint-Embedding Predictive Architecture (JEPA), to monitor the learning of predictive representations. Extensive experiments on challenging benchmarks demonstrate the versatility of our method, showing that TimeCapsule can achieve state-of-the-art performance. The code is available at: \url{https://github.com/Luoauoa/TimeCapsule.git}.
\end{abstract}





\maketitle

\section{Introduction}
\label{intro}
Multivariate Time Series (MvTS) data is one of the most ubiquitous forms of naturally generated data in the temporal physical world. Forecasting future events, whether short-term or long-term, based on these collected historical data, can support critical human activities, including finance \cite{sezer2020financial}, traffic \cite{guo2020traffic}, and weather prediction \cite{karevan2020weather}. Moreover, it enables us to fundamentally explore the mechanisms underlying the world's operations \cite{bradley1999time}.

\begin{figure}[htp]
\subfigure[Same temporal patterns shared by different variates]{
\begin{minipage}[hbtp]{\linewidth}
\includegraphics[width=\linewidth]{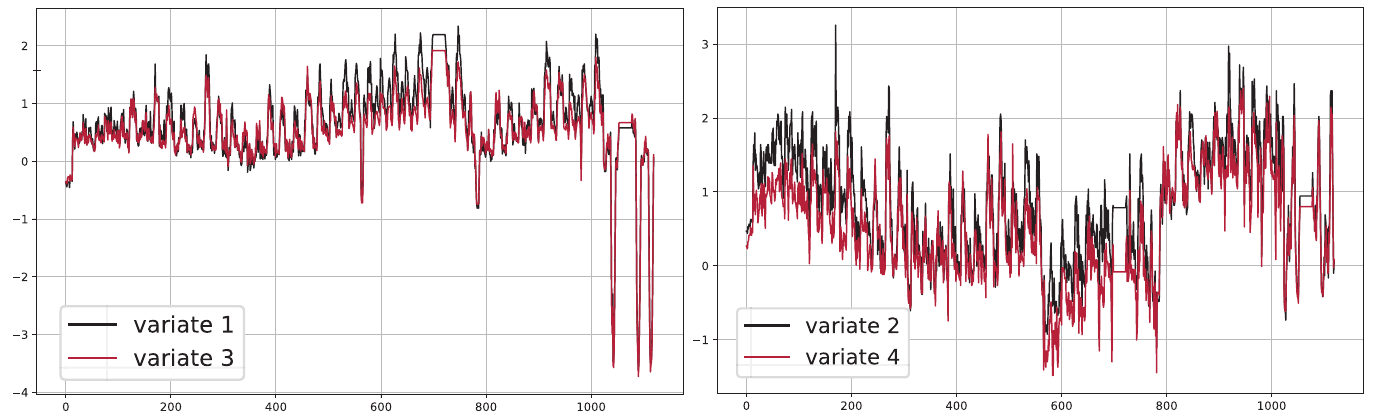}
\end{minipage}
}
\subfigure[Temporal pattern by sampling every 10 points]{
\begin{minipage}[hbtp]{\linewidth}
\includegraphics[width=\linewidth]{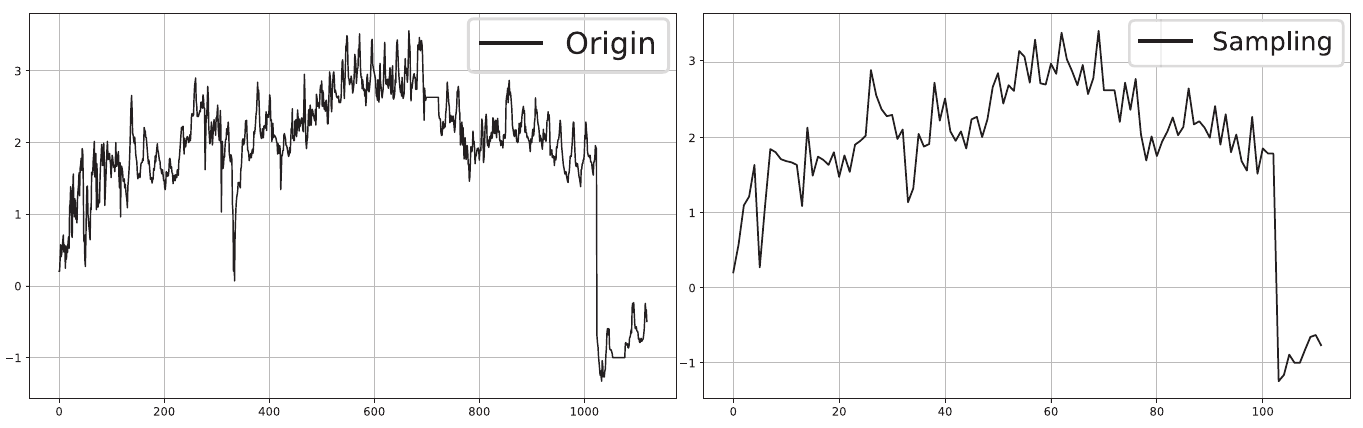}
\end{minipage}
}
\caption{We visualize the ETTh1 dataset as an example to illustrate the need to compress the information in different MvTS datasets' dimensions.}
\label{fig:0}
\end{figure}

With the rapid advancement of deep learning models, Long-term Time Series Forecasting (LTSF) has recently gained prominence. Unlike short-term forecasting, LTSF has traditionally posed significant challenges for classical statistical and machine learning methods, such as VAR, ARIMA, and random forests \cite{toda1994vector,box1970ari,kane2014randomf}. A series of groundbreaking works \cite{zhou2021informer, zeng2023dlinear,nie2022patch,liu2023itransformer} have been proposed to push the boundaries of this field, addressing limitations and advancing the community. Nevertheless, questions remain about how best to improve existing models, leaving room for highlighting potential synergies among existing methods and combining their strengths. For instance, Informer\cite{zhou2021informer} suggests that the learned attention map of transformers should be sparse and can be distilled into smaller representations. Models like FiLM \cite{zhou2022film} and FEDformer \cite{zhou2022fedformer} manipulate time series in compact frequency bases to capture key temporal correlations.
These advancements raise several compelling questions: Can these enhancements be conveniently extended to other dimensions? 
If so, a generalized transformation domain would be required.
Moreover, is time series data—or the information it contains—inherently compressible?
We can intuitively sense this point from two perspectives, as illustrated in Fig. \ref{fig:0}.

First, the various variates in a collected time series dataset often share very similar variation processes, suggesting that much of the information they carry is highly correlated, with minimal mutual informativeness. Therefore, adaptive recognition and distillation of complex intercorrelations within specific multivariate time series (MvTS) is fundamentally beneficial.
Second, predicting the long-term trend of a non-stationary time series is inherently challenging yet crucial. According to the sampling theorem, sparse sampling with larger intervals can still adequately represent the underlying trend. This insight suggests that encoding the long-term trend with fewer bits is indeed possible.

What's more, PatchTST applies the idea of patching to LTSF, enabling larger local receptive fields and greater efficiency. However, its reliance on predefined patch lengths introduces a rigidity that can be inflexible when dealing with varying input sequence lengths, thereby limiting its range of application. Additionally, this operation is not differentiable. Unlike image or text data, which contain explicit semantic information, splitting a time sequence into small chunks outside the training process may lead to unrecoverable information loss. 

While these methods individually have provided promising directions, it is shown that different time series may have different underlying preferences \cite{shao2024exploring}. Simplifying and combining these critical principles to build a versatile forecastor that can identify the intrinsic property of each dataset is appealing.

Recent discussions in the community have also drawn attention to the surprisingly strong performance of linear models when compared to transformer-based architectures in LTSF tasks.
It turns out that the effectiveness of transformers \cite{wang2022language, yuan2021vit} in LTSF remains inconsistent, often falling short of simpler linear models \cite{zeng2023dlinear}.
In this work, we propose that an effective architecture for LTSF should consist of two stages: predictive representation learning and generalized linear projection, in which the latter serves as an accurate predictor to learn generalized linear dependencies for the forecasting, while the former extracts abstract and informative representation from distinct datasets to improve forecasters' generality. Therefore, we propose TimeCapsule, a novel model employing a Chaining Bits Back with Asymmetric Numeral Systems (BB-ANS) \cite{townsend2019practical} like-architecture, which adopts weak transformer-based blocks as the encoder, while MLP-based blocks as the decoder, striking a balance between powerful representation learning and computational simplicity. Technically, our model is driven by three principles: 

\textbf{Multi-level Modeling}:  Multi-scale modeling stands out as an effective paradigm for improving LSTF performance \cite{ferreira2006multi}. Existing methods incorporate multiresolution analysis by up/downsampling through moving average and convolutional pooling layers \cite{challu2023nhits,wang2024timemixer}, or designing hierarchical structures \cite{liu2021pyraformer, zhang2023crossformer} to aggregate multi-scale features. Besides, time series decomposition is also a traditional and commonly used strategy to improve LTSF performance \cite{oreshkin2019nbits, zhou2022fedformer, wu2021autoformer}. In avoidance of complexity, we propose adding an extra dimension called level to the original time series, allowing the model to learn multi-level features within the representation space. This approach generalizes multi-scale learning and time series decomposition, making the learning process model-independent.

\textbf{Multi-mode Dependency}: iTransformer \cite{liu2023itransformer} has shown the benefits of leveraging correlations across dimensions other than time. However, focusing too heavily on non-temporal dimensions can risk neglecting important temporal dependencies, potentially degrading forecasting accuracy. To address this, we introduce a \underline{Mo}de-specific \underline{M}ulti-head \underline{S}elf-\underline{A}ttention (MoMSA) mechanism, leveraging tensor-based mode product techniques to capture dependencies along and across multiple dimensions, including temporal, variate, and level. 

\textbf{Compressed Representation Forecasting}: As illustrated above, to enable efficient information utilization, fast multi-mode attention computations, and better long-range history processing, we conclude that \textit{compression} is all we need. To be specific, rather than focusing exclusively on sparse attention and redundancy reduction, we employ low-rank transforms to replace patching, reducing dimensionality ahead of the attention computation, thus leading to an economic way to employ transformers. This strategy compresses the representation space, enabling efficient computation and robust long-range forecasting. The compressed space can also serve as the intermediate stage for learning forecasts, allowing us to map the representation into the future landscape and recover it back into the real temporal domain. 

Through this framework, we hope the neural network not only purifies redundant information but also learns to decompose, analyze, and recreate the underlying mechanisms of diverse time series. Our contributions can be summarized as follows: 
\begin{itemize}
\item We propose simple yet effective generalizations to existing LTSF techniques, including a Mode-Specific Multihead Self-Attention mechanism, resulting in a versatile forecasting model capable of handling diverse data characteristics.
\item We introduce JEPA into time series forecasting, making it a useful tool for monitoring and analyzing the process of predictive representation learning.
\item Extensive experiments on real-world datasets demonstrate the superiority of our approach, identifying areas for further exploration. 
\end{itemize}

\section{Related Works}
\subsection{Long-term Time Series modeling}
\begin{figure}[htbp]
\centering
  \includegraphics[scale=0.25]{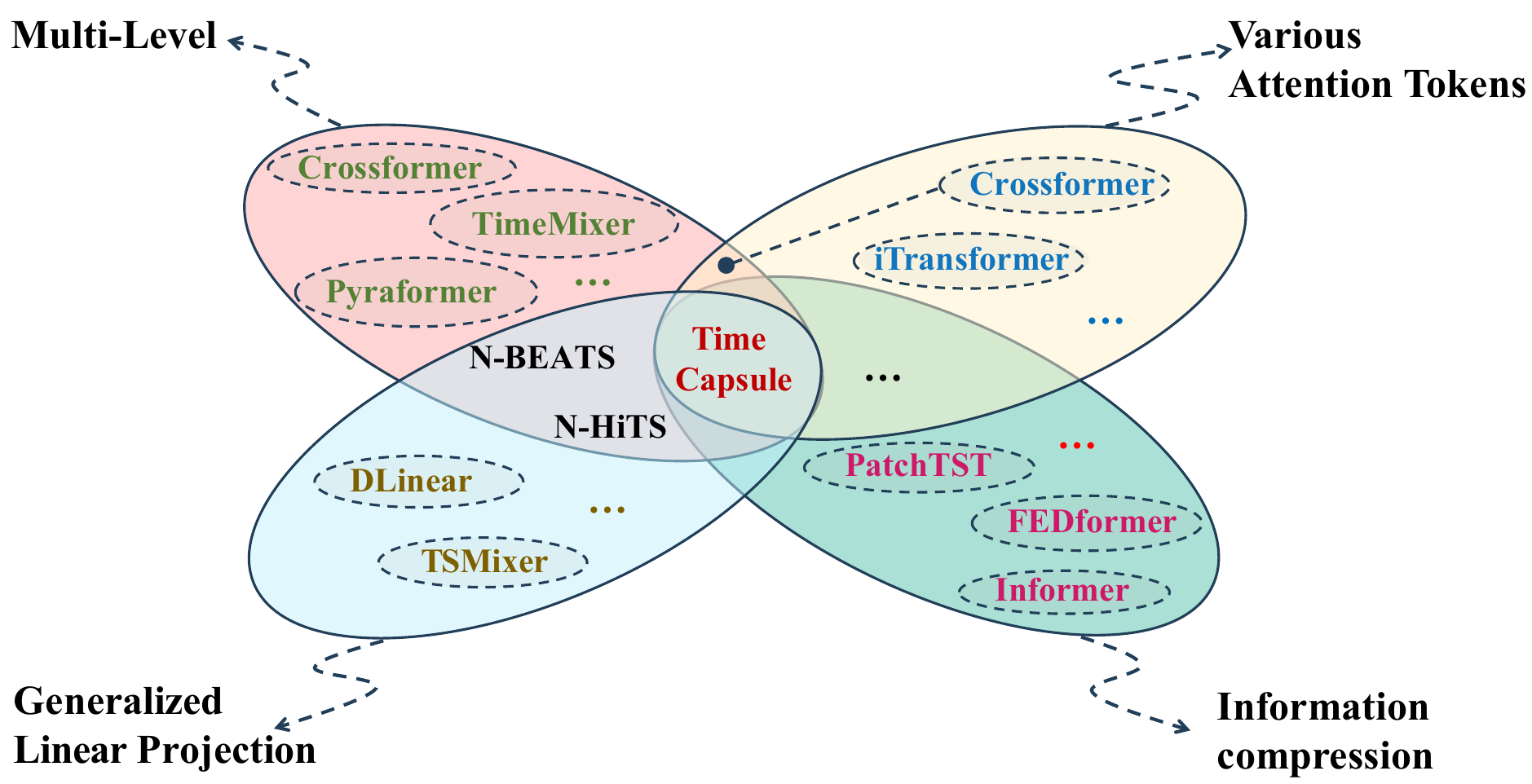}
  \caption{Categorization of advanced LTSF models into four groups based on their core techniques.}
  \label{fig:cata}
\end{figure}

Given the advantage of capturing long-term dependencies in long sequence data, researchers have increasingly applied transformers to LTSF tasks \cite{wen2022transformers}. Most recently, several significant works have been proposed to improve transformers' forecasting performance or reduce their computational complexity. In addition, some models that diverge from transformer architectures have also exhibited considerable promise in this area. Based on the strategies by which they achieve success, we can categorize these models into four groups, with potential overlaps, as presented in Fig. \ref{fig:cata}.

The $\textbf{first group}$ (e.g., Autoformer \cite{wu2021autoformer}, N-BEATs \cite{oreshkin2019nbits}, Pyraformer \citep{liu2021pyraformer}, N-Hits \cite{challu2023nhits}, Crossformer \cite{zhang2023crossformer}, TimeMixer \cite{wang2024timemixer}) focuses on multi-level modeling, which incorporates multiresolution/multi-scale analysis and series decomposition within the model. These techniques enable the model to learn both coarse and fine-grained features within time series, facilitating the capture of hierarchical temporal patterns. The $\textbf{second group}$ (e.g., Informer \cite{zhou2021informer}, FEDformer \cite{zhou2022fedformer}, PatchTST \cite{nie2022patch}), leverages information redundancy to filter and extract high-energy temporal correlations, or models time series in the form of temporal patches, avoiding the inclusion of finer, noisier information in the forecasting process, improving the model’s robustness and prediction accuracy., These advancements lead to progressive improvements in both effectiveness and efficiency. The $\textbf{third group}$ (e.g., Crossformer, iTransformer \cite{liu2023itransformer}) explores the impact of attention applied across various dimensions of time series, offering novel perspectives for time series correlation extraction. This group demonstrates that capturing dependencies across different modes can significantly enhance forecasting performance by providing a richer understanding of temporal and variable relationships. The $\textbf{fourth group}$ (e.g., DLinear \cite{zeng2023dlinear}, N-BEATs, N-Hits, TsMixer \cite{chen2023tsmixer}, TimeMixer) emphasizes the importance of generalized linear dependency modeling and capitalizes on linear or MLP-based architectures to establish highly effective forecasters. This validates the crucial role of learning the appropriate coefficients to combine the captured base components of time series and the dependencies they contain.

These four groups highlight four key factors in answering the question of how to model and forecast time series effectively from different aspects. However, each model typically focuses on building complex modules that address only parts of these factors. In contrast, our model integrates all of these interesting factors into a comprehensive yet streamlined design.

\subsection{Joint-Embedding Predictive Architecture} JEPA \cite{lecun2022path} realizes representation learning by optimizing an energy function between predicted representations of inputs and targets. It does not rely on explicit contrastive losses \cite{khosla2020supercontras}, but instead creates compatible embeddings through prediction, enhancing the flexibility and efficiency of representation learning. While JEPA has demonstrated success in learning predictable representations in vision tasks \cite{assran2023ijepa} and video tasks \cite{bardes2024vjepa}, these domains often benefit from spatial locality and semantic coherence, which are different from the temporal dependencies and complex long-range patterns present in time series data. Such distinctions pose unique challenges when applying JEPA to LTSF. Although some recent works, such as LaT-PFN \cite{verdenius2024lat} and TS-JEPA \cite{girgis2024tsjepa}, have explored JEPA in time series tasks, their focus differs significantly from the predictive demands of LTSF. Specifically, the former employs JEPA to construct a time series foundation model, demonstrating that JEPA can reinforce the latent embedding space of time series learning and result in superior zero-shot performance. In contrast, the latter, TS-JEPA, utilizes JEPA to facilitate the realization and enhancement of the effectiveness of semantic communication systems.

\section{Preliminaries}
\label{prepare}
We aim to predict the next $T_y$ steps of the time sequence, denoted as $Y=\{\large{y}_{\small{t_x}+1},\large{y}_{\small{t_x}+2},\cdots,\large{y}_{\small{t_x}+\small{t_y}}\}\in \mathbb{R}^{v\times t_y}$, based on the observed sequence $X=\{\large{x}_1,\large{x}_2,\cdots,\large{x}_{\small{t_x}}\}\in \mathbb{R}^{v\times t_x}$, where $v$ represents the number of variates and $t_x$ denotes the length of the input sequence. The key distinction in this work is that we treat MvTS as 3D tensor data, allowing the data itself to handle multi-scale modeling and series decomposition by introducing an additional dimension. Specifically, we extend the input from $\text{X} \in \mathbb{R}^{v \times t_x}$ to $\mathcal{X} \in \mathbb{R}^{v \times t_x \times 1}$. Throughout this paper, we use T, V, and L to represent the temporal dimension, variate dimension, and level dimension, respectively, with $t$, $v$, and $l$ indicating their corresponding lengths. We define $1 \leq t_c \ll t$, $1 \leq v_c \ll v$, and $l_c \geq 1$ as the compressed lengths of the respective dimensions. Additionally, $d$ denotes the size of the embedding space.

Mode production is a common arithmetic operation in tensor methods \citep{kernfeld2015tensor}, which relies on two fundamental concepts: tensor folding and tensor unfolding. For simplicity and easy understanding, we provide an \textbf{informal definition} of the mode product of a 3D tensor as follows,
\begin{definition}[Mode Product] 
Given a real 3D tensor $\mathcal{X} \in \mathbb{R}^{n_1\times n_2\times n_3}$, the result of mode-$3$ unfolding of $\mathcal{X}$ is the matrix $\mathbf{X}_{(3)}\in\mathbb{R}^{n_3\times n_1n_2}$, denoted by
\begin{equation*}
    \text{Fold}_{(3)}(\mathcal{X})=\mathbf{X}_{(3)}
\end{equation*}
and the mode-$3$ folding operation recovers the matrix back into the tensor, denoted by
\begin{equation*}
    \text{Unfold}(\mathbf{X}_{(3)})=\mathcal{X}
\end{equation*}
We then define the mode-$3$ production as
\begin{equation*}
\mathcal{X}\times_{3} \textbf{M}=\text{Unfold}(\textbf{M}\textbf{X}_{(3)})\in \mathbb{R}^{n_1\times n_2\times m}
\end{equation*}
where $\textbf{M}\in \mathbb{R}^{m\times n_3}$ is the transform factor. This production can be generalized to any mode of any tensor, leading to the definition of the mode-$k$ product.
\end{definition}

\section{Proposed Methodology}
\begin{figure*}[htbp]
\centering
  \includegraphics[width=\linewidth]{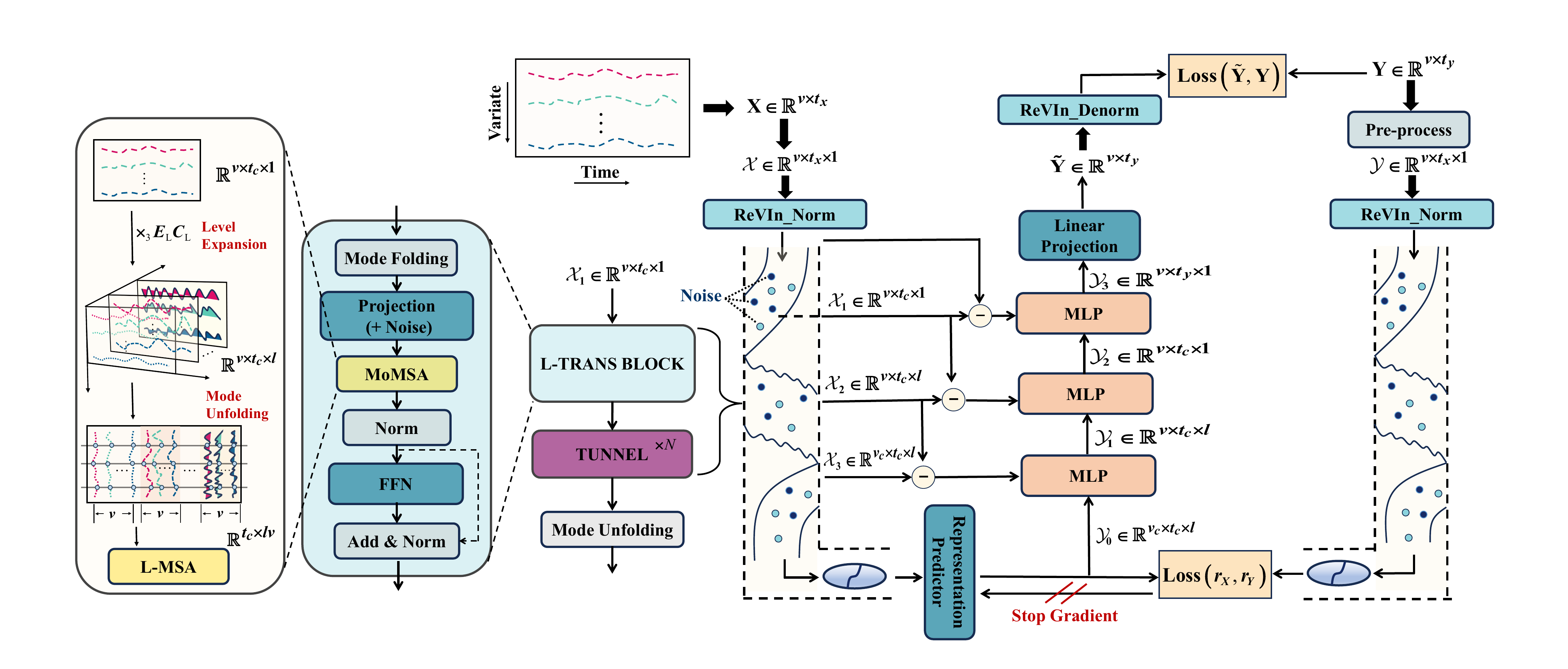}
  \caption{Overview of the TimeCapsule model. The original time series is transformed into a 3D representation by adding a level dimension, traversing through TransBlocks and tunnels (vanilla transformer blocks), and is then projected into the predictive space using JEPA. The compressed capsule is gradually recovered back into the real temporal domain.}
  \label{fig:2}
\end{figure*}
\subsection{Feed Forward Process}
\label{feedforward}
Generally, as depicted in Fig. \ref{fig:2}, TimeCapsule follows an asymmetric two-stage learning process: deep representation encoding and compressed information-based prediction. Internally, the encoder consists of three distinct stacks, each containing a transformer block followed by a series of tunnels. We place the time dimension T as the first block because it has a relatively long length, which should be compressed first to reduce the overall computational cost. The level expansion is placed second to enable multi-level learning in the representation space as efficiently as possible. Finally, we process the variable dimension. In contrast, the decoder is simpler, comprising just three MLP blocks. We will first present an overview of the forward process, then go over a detailed explanation of how each of the key components is built and operates within this framework.

The encoding part can be formulated as 
\begin{gather}
\mathcal{X} \leftarrow \text{X}\in\mathbb{R}^{v\times t_x} ,~~\mathcal{X}_0=\text{RevIn}(\mathcal{X})\in\mathbb{R}^{v\times t_x\times 1}\\
\mathcal{X}_1=\text{Tunnel}(\text{T-TransBlock}(\mathcal{X}_0))\in \mathbb{R}^{v\times t_c\times1}\\
\mathcal{X}_2=\text{Tunnel}(\text{L-TransBlock}(\mathcal{X}_1))\in \mathbb{R}^{v\times t_c\times l}\\
\mathcal{X}_3=\text{Tunnel}(\text{V-TransBlock}(\mathcal{X}_2))\in \mathbb{R}^{v_c\times t_c\times l}
\end{gather}
where $\text{RevIn}(\cdot)$ denotes the reversible instance normalization proposed by \citep{kim2021reversible}, and $\text{Tunnels}$ and $\text{TransBlocks}$ are all transformer-based blocks. The prefix of $\text{TransBlock}$ indicates the dimension along which the blocks are applied  (T for temporal, L for level, and V for variate). 

Next, the decoder operates in the reverse order of the encoding process:
\begin{gather}
\mathcal{Y}_0=\text{Repre\_Predictor}(\mathcal{X}_3)\in\mathbb{R}^{v_c\times t_c\times l} \label{eq:5}\\ 
\mathcal{Y}_1=\text{MLP(Cat}(\mathcal{Y}_0, \mathcal{B}_3))\in\mathbb{R}^{v\times t_c\times l} \label{eq:6}\\
\mathcal{Y}_2=\text{MLP(Cat}(\mathcal{Y}_1, \mathcal{B}_2))\in\mathbb{R}^{v\times t_c\times 1} \label{eq:7}\\
\mathcal{Y}_3=\text{MLP(Cat}(\mathcal{Y}_2, \mathcal{B}_1))\in\mathbb{R}^{v\times t_x\times 1} \label{eq:8}
\end{gather}
where $\text{Repre-Predictor}(\cdot)$  is a single linear layer that projects the deep representation into the future landscape, and $\text{Cat}(\cdot, \cdot)$ represents concatenation. $\mathcal{B}_1$, $\mathcal{B}_2$, $\mathcal{B}_3$ denote the residual information, which will be explained in detail later. The $\text{MLP}$ block contains three linear layers with an intermediate $\text{GELU}$ activation. Finally, we obtain the prediction result by another linear projection $\mathbb{R}^{t_y}\rightarrow\mathbb{R}^{t_y}$ and the inverse instance normalization.
\begin{gather}
\mathcal{Y} = \text{Proj}(\mathcal{Y}_3)\in \mathbb{R}^{v\times t_y\times 1},~~\mathcal{Y} \rightarrow \tilde{\text{Y}}\in \mathbb{R}^{v\times t_y} \\
\text{Y}=\text{RevIn}(\tilde{\text{Y}}) \in\mathbb{R}^{v\times t_y}
\end{gather}

\subsection{Main Components}
\subsubsection{Mode Specific Multi-head Self-Attention (MoMSA)}
MoMSA is designed to achieve two primary objectives: (a) to extend the ideas of crossformer \cite{zhang2023crossformer} and iTransformer \cite{liu2023itransformer} by forming MvTS tokens from a multi-mode view and abstracting dependencies along each mode; and (b) to maintain the same volume of information while shortening the length of each dimension, thereby reducing the overall computational cost of multi-mode self-attention. 

To accomplish this, we introduce the mode-$k$ product (see chapter \ref{prepare} for the definition). For instance, given an MvTS tensor $\mathcal{A}\in \mathbb{R}^{v_a\times t_a\times l_a}$ and a transform factor $\text{M} \in\mathbb{R}^{m\times t_a}$, MoMSA regarding to the temporal dimension operates as follows:
\begin{align}
\hat{\mathcal{A}} = \mathcal{A}\times_2\text{M},~~ \text{A}_1=\text{T-MSA}(\hat{\text{A}}_{(2)}) \in \mathbb{R}^{m\times v_a l_a}
\end{align}
where $\times_2$ denotes the mode-2 product, which represents a matrix multiplication along the second dimension of a 3D tensor. $\text{T-MSA}$ denotes the vanilla multi-head self-attention \citep{vaswani2017attention} applied to the $\text{T}$ (temporal) dimention. It is notable that with the setting of $m \le v_a$, the dimension can be compressed to an arbitrary length, resulting in a compressed attention map. 

When applying this procedure to the real case, we shall obtain $\mathcal{X}_1 \in \mathbb{R}^{v\times t_c\times 1}$ from the first pipe as shown in Fig. \ ref {fig:2}. However, it is risky to make such a compression on the original MvTS information. To address this, we take a number of protective measures within our TransBlock. In particular, before applying MoMSA, we project the information into the embedding space $\mathbb{R}^{t_x\times d}$ and introduce Gaussian noise at the start of the block, which may improve robustness during compression (channel coding). Furthermore, we set the transform factor as the product of two matrices
$$
\text{M}_\text{T}=\text{C}_\text{T}\text{E}_\text{T}
$$
where $\text{E}_\text{T}\in \mathbb{R}^{t_e\times t_x}$ extends the dimension and $\text{C}_\text{T}\in \mathbb{R}^{t_c\times t_e}$ does the compression, with $t_e\ge t_x \gg t_c$. By doing so, we enhance the information before compression, akin to the strategy used in Informer \cite{zhou2021informer} and FEDformer \cite{zhou2022fedformer}, but within a generalized tranform domain. It is noteworthy that the transform factor is not necessarily invertible, as we allow lossy compression to help reduce information redundancy. In a nutshell, the derivation of the temporal MoMSA can be formulated as 
\begin{equation}{
\label{momsa}
\text{MoMSA}(\mathcal{X}_0)=\text{T}\text{-MSA}((\text{Proj}(\mathcal{X}_0)+\mathcal{N}(0,1))\times_2\text{C}_\text{T}\text{E}_\text{T})
}\end{equation}
where $\text{T}$ is selected from the mode set $\{\text{T},\text{L}, \text{V}\}$ to represent the $\text{T-TransBlock}$. Subsequently, additional transforms shall be applied in a sequential manner, whereby the symbol T shall be replaced with an alternative mode and the corresponding mode product shall be engaged. This process shall ultimately result in the generation of a compressed 3D representation, i.e., 'time capsule', at the end of the encoder.

Interestingly, our MoMSA can be viewed as a folded patch-wise self-attention mechanism that abstracts inter-correlations across different representation spaces. As illustrated in the leftmost example of Fig. \ref{fig:2}, the two-dimensional information is obtained by folding the 3D tensor along the temporal dimension. This results in the formation of $l$ distinct groups, each representing a unique level, and containing $v$ variables. The attention token within the L-TransBlock is constituted by a combination of variables from different levels at each compressed timestamp. In this manner, traversing three TransBlocks with respect to various dimensions allows for the thorough capture of multi-mode dependencies. A comprehensive visual analysis can be found in the Appendix \ref{learn}.

\subsubsection{Residual Information Back} One crucial aspect of our model to ensure more accurate predictions lies in leveraging complete but filtered history information. Thus, it is required to compensate for the lost information during the decoding process. We transfer to the decoder the information calculated via residual subtractions instead of using the original $\mathcal{X}$, which may provide a shortcut for information retrieval.
\begin{gather}
\mathcal{B}_1 = (\mathcal{X}_0-\mathcal{X}_1\times_2\text{C}_\text{T}\text{E}_\text{T}) \in \mathbb{R}^{v\times t_x\times1}\\
\mathcal{B}_2 = (\mathcal{X}_1-\mathcal{X}_2\times_3\text{C}_\text{L}\text{E}_\text{L}) \in \mathbb{R}^{v\times t_c\times1}\\
\mathcal{B}_3 = (\mathcal{X}_2-\mathcal{X}_3\times_1\text{C}_\text{V}\text{E}_\text{V}) \in \mathbb{R}^{v\times t_c\times l}
\end{gather}
these residuals are subsequently used as outlined in Eq.\ref{eq:6} to Eq.\ref{eq:8}.

\subsubsection{Representation Prediction with JEPA Loss}
In addition to the time-domain prediction, the structure of TimeCapsule introduces inner predictions for the compressed representation. We hypothesize a potential gap (e.g., inter-space distribution shift (Dish-TS)) between the historical and future series. We argue that the use of reversible instance normalization (RevIn) \cite{kim2021reversible} does not affect the stationarity of the time series. The primary objective of RevIn is to mitigate the distribution shifts between the training and testing data. However, because the statistics it uses are calculated along the temporal dimension, the shape and stochastic properties of the series remain invariant.

A key challenge is how to track and validate this inner prediction. To address this, we employ the Joint-Embedding Predictive Architecture (JEPA), which not only offers a measure to evaluate the distance between observations and predictions in non-stationary, time-varying processes but also enables efficient contrastive loss in the representation space. This helps the internal prediction to converge and, in some cases, may improve the real prediction as well.
\begin{table*}[h]
\caption{Full results of multivariate forecasting. For TimeCapsule, we use superscript $\dagger$ to denote the employment of JEPA training. The lookback length $T = 96$ and prediction lengths $S \in \{24, 36, 48, 60\}$ for ILI, $S \in \{96, 192, 336, 720\}$ and fixed lookback length $T = 512$ for others; For other models, lookback lengths are searched for the best performance.} 
\label{tab:1}
\centering
\resizebox{\textwidth}{!}{\begin{tabular}{c|c|cc|cc|cc|cc|cc|cc|cc|cc|cc|cc}
\toprule
\multicolumn{2}{c}{Models} & \multicolumn{2}{c|}{\textbf{TimeCapsule$^\dagger$}} & \multicolumn{2}{c|}{\textbf{TimeCapsule}} & \multicolumn{2}{c|}{iTransformer (\citeyear{liu2023itransformer})} & \multicolumn{2}{c|}{TimeMixer (\citeyear{wang2024timemixer})} & \multicolumn{2}{c|}{PatchTST (\citeyear{nie2022patch})} & \multicolumn{2}{c|}{Crossformer (\citeyear{zhang2023crossformer})} & \multicolumn{2}{c|}{DLinear (\citeyear{zeng2023dlinear})} & \multicolumn{2}{c|}{TimesNet (\citeyear{wu2022timesnet})} & \multicolumn{2}{c|}{FEDformer (\citeyear{zhou2022fedformer})} & \multicolumn{2}{c}{Informer (\citeyear{zhou2021informer})} \\
\cmidrule{3-22} 
\multicolumn{2}{c}{Metric} & MSE & MAE & MSE & MAE & MSE & MAE & MSE & MAE & MSE & MAE & MSE & MAE & MSE & MAE & MSE & MAE & MSE & MAE & MSE & MAE \\ \toprule
\multirow{4}{*}{PEMS04} & 96 & \textbf{0.099} & \textbf{0.202} & \underline{0.110} & \underline{0.211} & 0.164 & 0.280 & 0.122 & 0.229 & 0.161 & 0.280 & 0.112 & 0.224 & 0.196 & 0.296 & 0.159 & 0.266 & 0.573 & 0.565 & 0.189 & 0.304 \\
 & 192 & \textbf{0.117} & \textbf{0.222} & \underline{0.127} & \underline{0.224} & 0.216 & 0.316 & 0.141 & 0.239 & 0.178 & 0.290 & 0.134 & 0.236 & 0.213 & 0.310 & 0.179 & 0.282 & 0.655& 0.624 & 0.229 & 0.335 \\
& 336 & \textbf{0.126} & \textbf{0.229} & \underline{0.133} & \underline{0.230} & 0.189 & 0.288 & 0.153 & 0.254 & 0.193 & 0.302 & 0.190 & 0.286 & 0.235 & 0.327 & 0.169 & 0.269 & 1.365 & 0.920 & 0.217 & 0.323 \\
 & 720 & \textbf{0.137} & \textbf{0.239} & 0.187 & 0.285 & 0.251 & 0.351 & \underline{0.174} & \underline{0.276} & 0.233 & 0.338 & 0.235 & 0.331 & 0.327 & 0.395 & 0.187 & 0.286 & 0.873 & 0.728 & 0.310 & 0.391 \\
 \midrule
 \multirow{4}{*}{Weather} & 96 & \textbf{0.141} & \textbf{0.186} & \underline{0.142} & \underline{0.188} & 0.159 & 0.208 & 0.147 & 0.198 & 0.149  & 0.196 & 0.146 & 0.212 & 0.170 & 0.230 & 0.170 & 0.219 & 0.223 & 0.292 & 0.218 & 0.255  \\
 & 192 & \textbf{0.187} & \textbf{0.232} & \underline{0.188} & \underline{0.235} & 0.200 & 0.248 & 0.192 & 0.243 & 0.193 & 0.240 & 0.195 & 0.261 & 0.212 & 0.267 & 0.222 & 0.264 & 0.252 & 0.322 & 0.269 & 0.306 \\
 & 336 & \textbf{0.239} & \textbf{0.272} & \underline{0.241} & \underline{0.274} & 0.253 & 0.289 & 0.247 & 0.284 & 0.244 & 0.281 & 0.268 & 0.325 & 0.257 & 0.305 & 0.293 & 0.310 & 0.327 & 0.371 & 0.320 & 0.340 \\
 & 720 & \textbf{0.309} & \textbf{0.323} & \underline{0.311} & \underline{0.324} & 0.321 & 0.338 & 0.318 & 0.330 & 0.314 & 0.332 & 0.330 & 0.380 & 0.318 & 0.356 & 0.360 & 0.355 & 0.424 & 0.419 & 0.392 & 0.390 \\
 \midrule
\multirow{4}{*}{Traffic} & 96 & \underline{0.361} & \underline{0.246} & \textbf{0.355} & \textbf{0.244} & 0.363 & 0.265 & 0.369 & 0.257 & 0.370 & 0.262 & 0.514 & 0.282 & 0.410 & 0.282 & 0.600 & 0.313 & 0.593 & 0.365 & 0.664 & 0.371 \\
 & 192 & \underline{0.383} & \underline{0.257} & \textbf{0.378} & \textbf{0.256} & 0.385 & 0.273 & 0.400 & 0.272 & 0.386 & 0.269 & 0.501 & 0.273 & 0.423 & 0.288 & 0.619 & 0.328 & 0.614 & 0.375 & 0.724 & 0.396 \\
 & 336 & \underline{0.393} & \underline{0.262} & \textbf{0.390} & \textbf{0.262} & 0.396 & 0.277 & 0.407 & 0.272 & 0.396 & 0.275 & 0.507 & 0.278 & 0.436 & 0.296 & 0.627 & 0.330 & 0.609 & 0.373 & 0.796 & 0.435 \\
 & 720 & \underline{0.430} & \textbf{0.282} & \textbf{0.429} & \underline{0.282} & 0.445 & 0.312 & 0.461 & 0.316 & 0.435 & 0.295 & 0.571 & 0.301 & 0.466 & 0.315 & 0.659 & 0.342 & 0.646 & 0.394 & 0.823 & 0.453 \\
 \midrule
 \multirow{4}{*}{Electricity} & 96 & \textbf{0.125} & \textbf{0.218} & \underline{0.126} & \underline{0.219} & 0.138 & 0.237 & 0.131 & 0.224 & 0.133 & 0.233 & 0.135 & 0.237 & 0.140 & 0.237 & 0.164 & 0.267 & 0.186 & 0.302 & 0.214 & 0.321 \\
 & 192 & \textbf{0.146} & \textbf{0.238} & \underline{0.149} & \underline{0.242} & 0.157 & 0.256 & 0.151 & 0.242 & 0.150 & 0.248 & 0.160 & 0.262 & 0.154 & 0.250 & 0.180 & 0.280 & 0.201 & 0.315  & 0.245 & 0.350 \\
 & 336 & \textbf{0.158} & \textbf{0.255} & 0.171 & 0.269 & \underline{0.167} & 0.264 & 0.169 & \underline{0.260} & 0.168 & 0.267 & 0.182 & 0.282 & 0.169 & 0.268 & 0.190 & 0.292 & 0.218 & 0.330 & 0.294 & 0.393 \\
 & 720 & \textbf{0.187} & \textbf{0.280} & \underline{0.194} & 0.287 & 0.194 & \underline{0.286} & 0.227 & 0.312 & 0.202 & 0.295 & 0.246 & 0.337 & 0.204 & 0.301 & 0.209 & 0.307 & 0.241 & 0.350 & 0.306 & 0.393 \\
 \midrule
\multirow{4}{*}{ILI} & 24 & \textbf{1.675} & \textbf{0.793} & 3.115 & 1.110 & \underline{1.783} & 0.846 & 1.807 & \underline{0.820} & 1.840 & 0.835 & 2.981 & 1.096 & 2.208 & 1.031 & 2.009 & 0.926 & 2.400 & 1.020 & 2.738 & 1.151 \\
 & 36 & \textbf{1.619} & \textbf{0.796} & 1.740 & \underline{0.841} & 1.746 & 0.860 & 1.896 & 0.927 & \underline{1.724} & 0.845 & 3.295 & 1.162 & 2.032 & 0.981 & 2.552 & 0.997 & 2.410 & 1.005 & 2.890 & 1.145 \\
 & 48 & \textbf{1.653} & \textbf{0.835} & \underline{1.682} & \underline{0.856} & 1.716 & 0.898 & 1.753 & 0.866 & 1.762 & 0.863 & 3.586 & 1.230 & 2.209 & 1.063 & 1.956 & 0.919 & 2.592 & 1.033 & 2.742 & 1.136 \\
 & 60 & \underline{1.653} & \underline{0.830} & \textbf{1.627} & \textbf{0.827} & 1.960 & 0.977 & 1.828 & 0.930 & 1.752 & 0.894 & 3.693 & 1.256 & 2.292 & 1.086 & 2.178 & 0.962 & 2.539 & 1.070 & 2.825 & 1.139 \\
 \midrule
\multirow{4}{*}{Solar} & 96 & 0.173 & \underline{0.229} & \textbf{0.170} & \textbf{0.225} & 0.188 & 0.242 & 0.178 & 0.231 & 0.170 & 0.234 & 0.183 & 0.230 & 0.216 & 0.287 & 0.285 & 0.330 & 0.509 & 0.530 & 0.338 & 0.373  \\
 & 192 & \textbf{0.188} & \underline{0.242} & \underline{0.189} & 0.245 & 0.201 & 0.259 & 0.209 & 0.273 & 0.204 & 0.302 & 0.208 & \textbf{0.226} & 0.244 & 0.305 & 0.309 & 0.342 & 0.474 & 0.500 & 0.375 & 0.391 \\
 & 336 & \underline{0.194} & \textbf{0.248} & 0.195 & \underline{0.248} & 0.195 & 0.259 & \textbf{0.190} & 0.256 & 0.212 & 0.293 & 0.203 & 0.260 & 0.263 & 0.319 & 0.335 & 0.365 & 0.338 & 0.439 & 0.417 & 0.416 \\
 & 720 & 0.204 & \textbf{0.254} & \textbf{0.203} & \underline{0.255} & 0.223 & 0.281 & \underline{0.203} & 0.261 & 0.217 & 0.307 & 0.215 & 0.256 & 0.264 & 0.324 & 0.346 & 0.355 & 0.365 & 0.459 & 0.390 & 0.407 \\
 \midrule
  \multirow{4}{*}{ETTm2} & 96 & \textbf{0.160} & \textbf{0.247} & \underline{0.161} & \underline{0.249} & 0.175 & 0.266 & 0.172 & 0.265 & 0.165 & 0.254 & 0.263 & 0.359 & 0.164 & 0.255 & 0.190 & 0.266 & 0.219 & 0.306 & 0.216 & 0.302 \\
 & 192 & \textbf{0.215} & \textbf{0.288} & \underline{0.216} & \underline{0.290} & 0.242 & 0.312 & 0.236 & 0.304 & 0.221 & 0.292 & 0.361 & 0.425 & 0.224 & 0.304 & 0.251 & 0.308 & 0.294 & 0.357 & 0.324 & 0.367 \\
 & 336 & \textbf{0.265} & \textbf{0.320} & \underline{0.268} & \underline{0.321} & 0.282 & 0.340 & 0.273 & 0.329 & 0.275 & 0.325 & 0.469 & 0.496 & 0.277 & 0.337 & 0.322 & 0.350 & 0.362 & 0.401 & 0.424 & 0.429 \\
 & 720 & \textbf{0.341} & \textbf{0.370} & \underline{0.346} & \underline{0.374} & 0.378 & 0.398 & 0.366 & 0.393 & 0.360 & 0.380 & 1.263 & 0.857 & 0.371 & 0.401 & 0.414 & 0.403 & 0.459 & 0.450 & 0.581 & 0.500 \\
 \midrule
 \multirow{4}{*}{ETTm1} & 96 & \textbf{0.284} & \textbf{0.341} & \underline{0.287} & \underline{0.343} & 0.300 & 0.353 & 0.336 & 0.371 & 0.290 & 0.343 & 0.310 & 0.361 & 0.299 & 0.343 & 0.377 & 0.398 & 0.467 & 0.465 & 0.430 & 0.424 \\
 & 192 & \textbf{0.325} & \textbf{0.364} & \underline{0.325} & 0.365 & 0.345 & 0.382 & 0.370 & 0.389 & 0.329 & 0.368 & 0.363 & 0.402 & 0.334 & \underline{0.364} & 0.405 & 0.411 & 0.610 & 0.524 & 0.550 & 0.479 \\
 & 336 & \textbf{0.354} & \textbf{0.382} & \underline{0.359} & 0.387 & 0.374 & 0.398 & 0.397 & 0.410 & 0.360 & 0.390 & 0.408 & 0.430 & 0.365 & \underline{0.384} & 0.443 & 0.437 & 0.618 & 0.544 & 0.654 & 0.529 \\
 & 720 & \textbf{0.413} & \underline{0.416} & \underline{0.413} & 0.417 & 0.429 & 0.430 & 0.463 & 0.446 & 0.416 & 0.422 & 0.777 & 0.637 & 0.418 & \textbf{0.415} & 0.495 & 0.464 & 0.615 & 0.551 & 0.714 & 0.578 \\
 \midrule
  \multirow{4}{*}{ETTh2} & 96 & \textbf{0.272} & \textbf{0.337} & 0.279 & 0.339 & 0.297 & 0.348 & 0.280 & 0.350 & \underline{0.275} & \underline{0.337} & 0.611 & 0.557 & 0.302 & 0.368 & 0.319 & 0.363 & 0.338 & 0.380 & 0.378 & 0.402  \\
 & 192 & \textbf{0.333} & \textbf{0.377} & \underline{0.336} & \underline{0.378} & 0.371 & 0.403 & 0.351 & 0.390 & 0.348 & 0.384 & 0.810 & 0.651 & 0.405 & 0.433 & 0.411 & 0.416 & 0.415 & 0.428 & 0.462 & 0.449  \\
 & 336 & 0.367 & 0.409 & \textbf{0.366} & \underline{0.406} & 0.404 & 0.428 & \underline{0.366} & 0.413 & 0.368 & \textbf{0.404} & 0.928 & 0.698 & 0.496 & 0.490 & 0.415 & 0.443 & 0.378 & 0.451 & 0.426 & 0.449  \\
 & 720 & \textbf{0.381} & \textbf{0.422} & \underline{0.382} & \underline{0.422} & 0.424 & 0.444 & 0.401 & 0.436 & 0.406 & 0.441 & 1.094 & 0.775 & 0.766 & 0.622 & 0.429 & 0.445 & 0.479 & 0.485 & 0.401 & 0.449  \\
   \midrule
  \multirow{4}{*}{ETTh1} & 96 & \textbf{0.362} & \textbf{0.394} & \underline{0.364} & \underline{0.396} & 0.386 & 0.405 & 0.373 & 0.401 & 0.376 & 0.396 & 0.411 & 0.435 & 0.379 & 0.403 & 0.389 & 0.412 & 0.379 & 0.419 & 0.709 & 0.563  \\
 & 192 & \textbf{0.399} & \textbf{0.418} & \underline{0.401} & \underline{0.420} & 0.424 & 0.440 & 0.415 & 0.425 & 0.409 & 0.425 & 0.409 & 0.438 & 0.404 & 0.413 & 0.440 & 0.443 & 0.419 & 0.443 & 0.724 & 0.570 \\
 & 336 & \textbf{0.424} & \textbf{0.432} & \underline{0.429} & \underline{0.435} & 0.449 & 0.460 & 0.438 & 0.450 & 0.431 & 0.444 & 0.433 & 0.457 & 0.440 & 0.440 & 0.482 & 0.465 & 0.455 & 0.464 & 0.732 & 0.581  \\
 & 720 & \textbf{0.424} & \textbf{0.447} & \underline{0.424} & \underline{0.448} & 0.495 & 0.487 & 0.486 & 0.484 & 0.457 & 0.477 & 0.501 & 0.514 & 0.469 & 0.489 & 0.525 & 0.501 & 0.474 & 0.488 & 0.760 & 0.616  \\
\midrule
\multicolumn{1}{c|}{1st Count} & & 31 & 32 & 8 & 5 & 0 & 0 & 1 & 0 & 0 & 1 & 0 & 1 & 0 & 1 & 0 & 0 & 0 & 0 & 0 & 0  \\
\bottomrule
\end{tabular}}
\end{table*}

To implement JEPA, another encoder is required to convert the target into the same representation space as the input. In line with the strategies used in I-JEPA and V-JEPA, we continue to obtain the target encoder by applying the Exponential Moving Average (EMA) of the input encoder. However, due to the nature of LTSF, it is problematic to deal with the inconsistency of the sequence length between the input and output. We disentangle this problem by preprocessing the target sequence to match the input length as follows,
\begin{equation}
\text{Y}=
\begin{cases}
   \text{Zero\_Padding}(\text{Y},t_x-t_y) & \text{if}~t_y<t_x\,\\
    \text{EMA}(\{\text{Y}_{(k-1)t_x+1:kt_x}\}_{k=1}^{\lceil t_y//t_x\rceil}) & \text{if}~t_y>t_x
\end{cases}
\end{equation}
i.e., in cases where $t_y < t_x$, we pad $\text{Y}$ with zeros; and when $t_y$ is larger, we can reasonably consider the whole series as a carrier of information, and the correlations between nearby timestamps are stronger. Under this assumption, we split the sequence $\text{Y}$ into subsequences of length $t_x$, then apply EMA to these chunks to form integrated time information. This enables an efficient computation of the predictive representation loss:
$$
\text{Loss}(\text{Enc}_x(\text{X}),~sg(\text{Enc}_y(\text{Y}))
$$
where $sg(\cdot)$ denotes the stop gradient operator. By incorporating this loss, we can quantify the distance between the learned representation and the target representation in the prediction space, and by default, we try to add this value to the final loss.
\section{Experiment}
To evaluate the performance and versatility of TimeCapsule, we conduct extensive experiments on ten diverse public datasets and compare the forecasting results against eight widely recognized forecasting models. More details on the datasets are provided in Appendix \ref{appen:a}. 

\subsection{Forecasting Results}
\subsubsection{Datasets and Baselines.}
We utilize datasets from various domains, including electricity (ETTh1, ETTh2, ETTm1, ETTm2, Electricity), environment (Weather), energy (Solar-Energy), transportation (PEMS04, Traffic), and health (ILI). The forecasting methods against which we compare include iTransformer \cite{liu2023itransformer}, TimeMixer \cite{wang2024timemixer}, PatchTST \cite{nie2022patch}, Crossformer \cite{zhang2023crossformer}, DLinear \cite{zeng2023dlinear}, TimesNet \cite{wu2022timesnet}, FEDformer \cite{zhou2022fedformer} and Informer \cite{zhou2021informer}.

\subsubsection{Main Configurations.}
All experiments are conducted on four NVIDIA 4090 GPUs with 24GB of memory each. To align with JEPA, we use AdamW \cite{loshchilov2017decoupled} as the optimizer and Huber loss \cite{meyer2021huber} as the default loss function. Results are obtained using the random seed 2021. The batch size for each case is selected within the range of 32 to 128, and the learning rate is determined through a grid search between 1e-4 and 2e-3. TimeCapsule consists of 0 to 2 blocks of tunnels, with the compression dimension length chosen from the set \{1, 4, 8, 32\}. To test the model’s ability to utilize long-range historical data, we use a lookback window of 512 for most datasets.

\subsubsection{Main Results.}
Given the numerous benchmarks proposed in this area and a recent hidden bug discovered in the testing phase codes, making fair and trustworthy comparisons under a unified setting has proven challenging.
Thanks to the contributions from TFB \cite{qiu2024tfb}, a comprehensive and reliable benchmark specifically designed for LTSF is now available, with results obtained through meticulous adjustments. To ensure objective comparisons, the experimental results reported in this section are partially derived from their work.
 
As shown in Table \ref{tab:1}, where the best results are highlighted in bold and the second best in underlined, each baseline model demonstrates distinct advantages across different scenarios. However, our model consistently achieves or approaches the best performance across all datasets and forecasting horizons. While improvements compared to the second-best results are sometimes marginal, TimeCapsule excels particularly in very long-term forecasting. These observations underscore the effectiveness and flexibility of our model, which integrates four key techniques in time series modeling. A central argument we propose is that different strategies may respond to various datasets to differing extents, revealing unique underlying characteristics. Taking the spatio-temporal dataset PEMS04 for a case study, TimeCapsule shows considerable progress compared to the second-best models, Crossformer and TimeMixer, with an average reduction of 15.8\% in MAE and 10\% in MSE. Notably, aside from Crossformer and TimeMixer, which emphasize careful multi-scale modeling, the performances of all other forecasters significantly degrade. This implies that forecasting on PEMS04 heavily relies on multi-scale modeling, a capability that TimeCapsule successfully possesses. Besides, our model outperforms PatchTST, particularly on large datasets, without using patching techniques. This indicates that TimeCapsule can also effectively leverage long historical information through dimensional compression and MLP use.

\subsection{Model Analysis}
In the following analyses, we keep the JEPA loss active throughout the entire training process, except in experiments where it is directly involved.

\begin{table}[htbp]
\caption{Ablation on the information compensation design. We remove the residual information connection or replace $\mathcal{B}$ with the original information $\mathcal{X}$. The resulting performance variations are then presented in terms of mean squared error (MSE).}
\label{tab:2}
\centering
\resizebox{.48\textwidth}{!}{\begin{tabular}{c|cccc|cccc|cccc}
\toprule
\multirow{3}{*}{Method} & \multicolumn{4}{c}{ETTh2} & \multicolumn{4}{c}{ETTm2} & \multicolumn{4}{c}{PEMS04}  \\
\cmidrule(r){2-5} \cmidrule(r){6-9} \cmidrule(r){10-13}  
& 96 & 192 & 336 & 720 & 96 & 192 & 336 & 720 & 96 & 192 & 336 & 720\\
\midrule
Origin & \textbf{0.272} & \textbf{0.333} & \textbf{0.367} & \textbf{0.381} & \textbf{0.160} & \textbf{0.215} & \textbf{0.265} & \textbf{0.341} & \textbf{0.099} & \textbf{0.117} & \textbf{0.126} & \textbf{0.137} \\
\midrule
Replace Info & 0.276 & 0.347 & 0.374 & 0.410 & 0.161 & 0.219 & 0.269 & 0.352 & 0.114 & 0.130 & 0.130 & 0.163 \\
\midrule
w/o Info Back & 0.371 & 0.387 & 0.393 & 0.462 & 0.289 & 0.317 & 0.352 & 0.422 & 0.273 & 0.247 & 0.290 & 0.339 \\
\bottomrule
\end{tabular}}
\end{table}
\subsubsection{Effects of residual information compensation.} Information compensation should be indispensable if it is reasonable to construct such a deep representation utilization strategy that adheres to the principle of lossy compression. To verify its effects, we compare model performance after removing the compensation and replacing it with the original information. The results, listed in Table \ref{tab:2}, reveal a dramatic drop in performance without paying back the lost information. This phenomenon indicates that the compressed information cannot be independently recovered under our default settings. However, when comparing our results with those of FEDformer and Informer, which also focus on redundancy reduction, our model demonstrates competitive results, validating the efficacy of our compression strategy. Additionally, compared to using the original information, the improvements observed on datasets like ETTh2 and ETTm2 are marginal. However, as the prediction horizon increases, the benefits of residual information become more pronounced. This suggests that redundancy reduction plays a more critical role in very long-term forecasting. For shorter forecasting lengths, the available information is typically sufficient. In contrast, when the forecasting horizon approaches or exceeds the input length, more compact and precise information is necessary, making the forecaster more sensitive to the quality and purity of the information provided. Notably, some degradations persist even with complete original information since it is sometimes challenging for our model to retrieve useful parts amid redundancy, which underscores the importance of our residual information computation.
\begin{figure}[h]
\subfigure[MLP (SAN) achieves $\text{MSE}=0.378$, this model explicitly predicts non-stationary statistics using SAN \cite{liu2024adaptivenorm}.]{
\begin{minipage}[hbtp]{\linewidth}
\includegraphics[width=\linewidth]{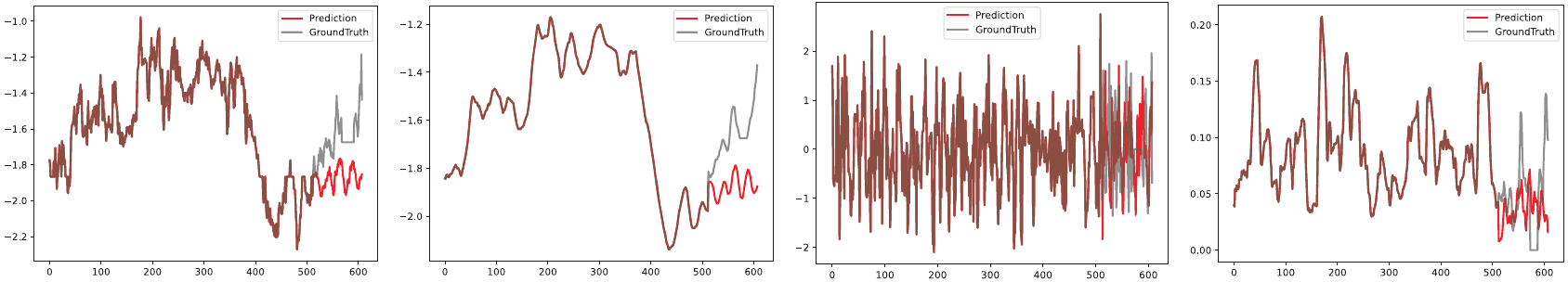}
\end{minipage}
}
\subfigure[iTransformer (version of non-stationary transformer) achieves $\text{MSE}=0.391$, this model explicitly models the non-stationary statistics within the non-stationary transformer \cite{liu2022nonsta}.]{
\begin{minipage}[hbtp]{\linewidth}
\includegraphics[width=\linewidth]{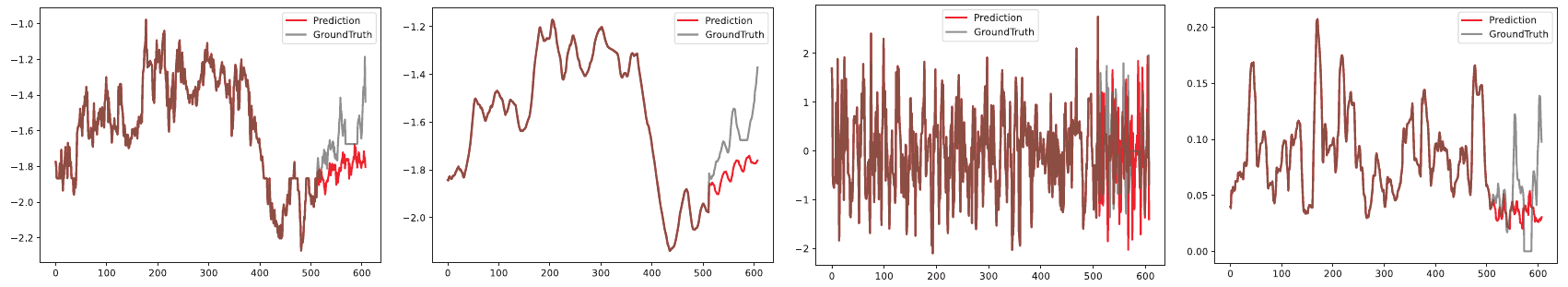}
\end{minipage}
}
\subfigure[MLP (RevIn) achieves $\text{MSE}=0.366$, this model is selected as a control group for the use of RevIn \cite{kim2021reversible}.]{
\begin{minipage}[hbtp]{\linewidth}
\includegraphics[width=\linewidth]{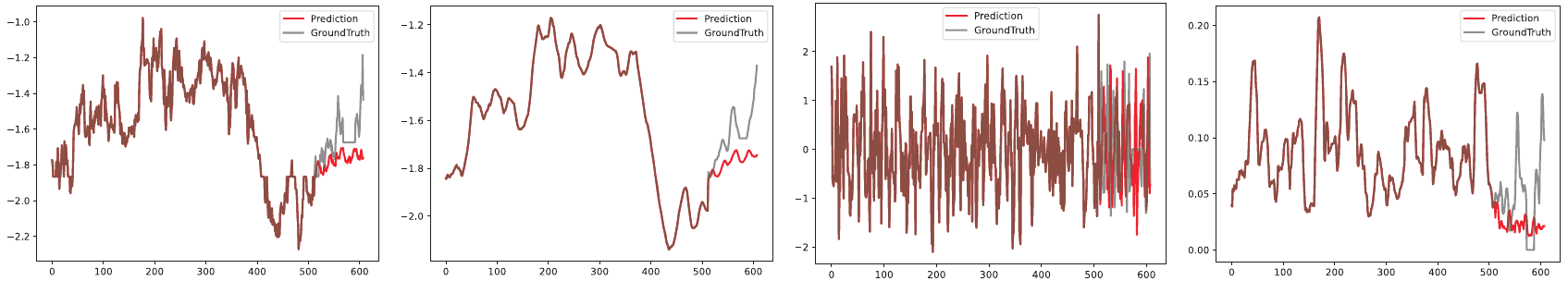}
\end{minipage}
}
\subfigure[TimeCapsule achieves $\text{MSE}=0.362$]{
\begin{minipage}[hbtp]{\linewidth}
\includegraphics[width=\linewidth]{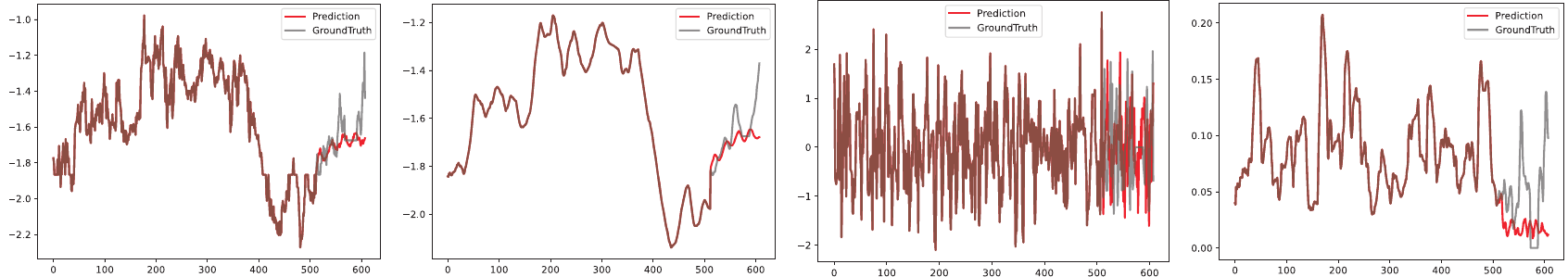}
\end{minipage}
}
\caption{We visualize ETTh1 dataset as a typical example to illustrate 512-96 predictions of different components (trend, stationary, deviation) of time series. All results are obtained under the recommended settings of the source code.}
\label{fig:y}
\end{figure}

\subsubsection{Effects of Internal Prediction and JEPA Utilization.}
\begin{figure}[htbp]
\subfigure{
\begin{minipage}[tb]{\linewidth}
\includegraphics[width=\linewidth]{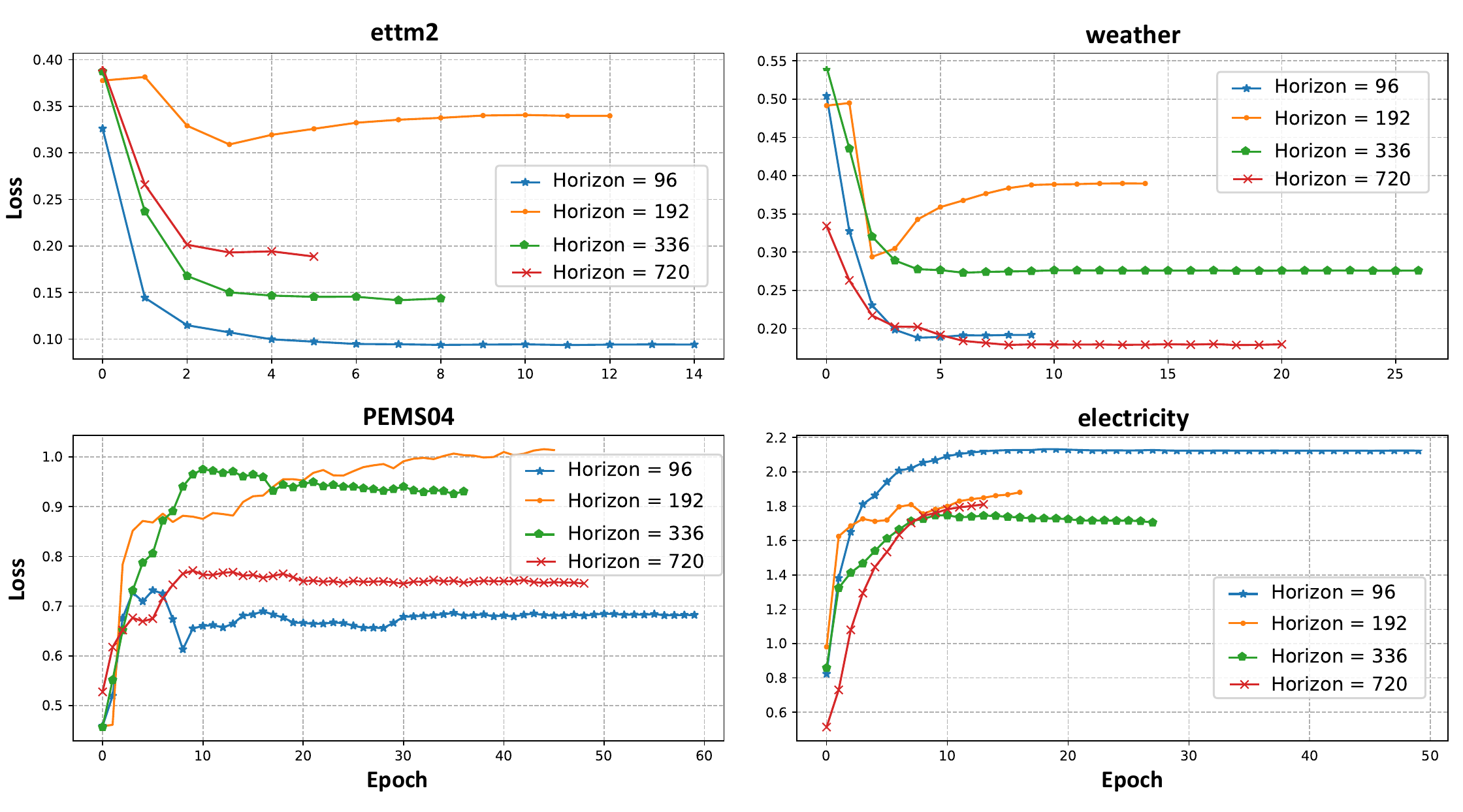}
\end{minipage}
}\\
\subfigure{
\begin{minipage}[tb]{\linewidth}
\includegraphics[width=\linewidth]{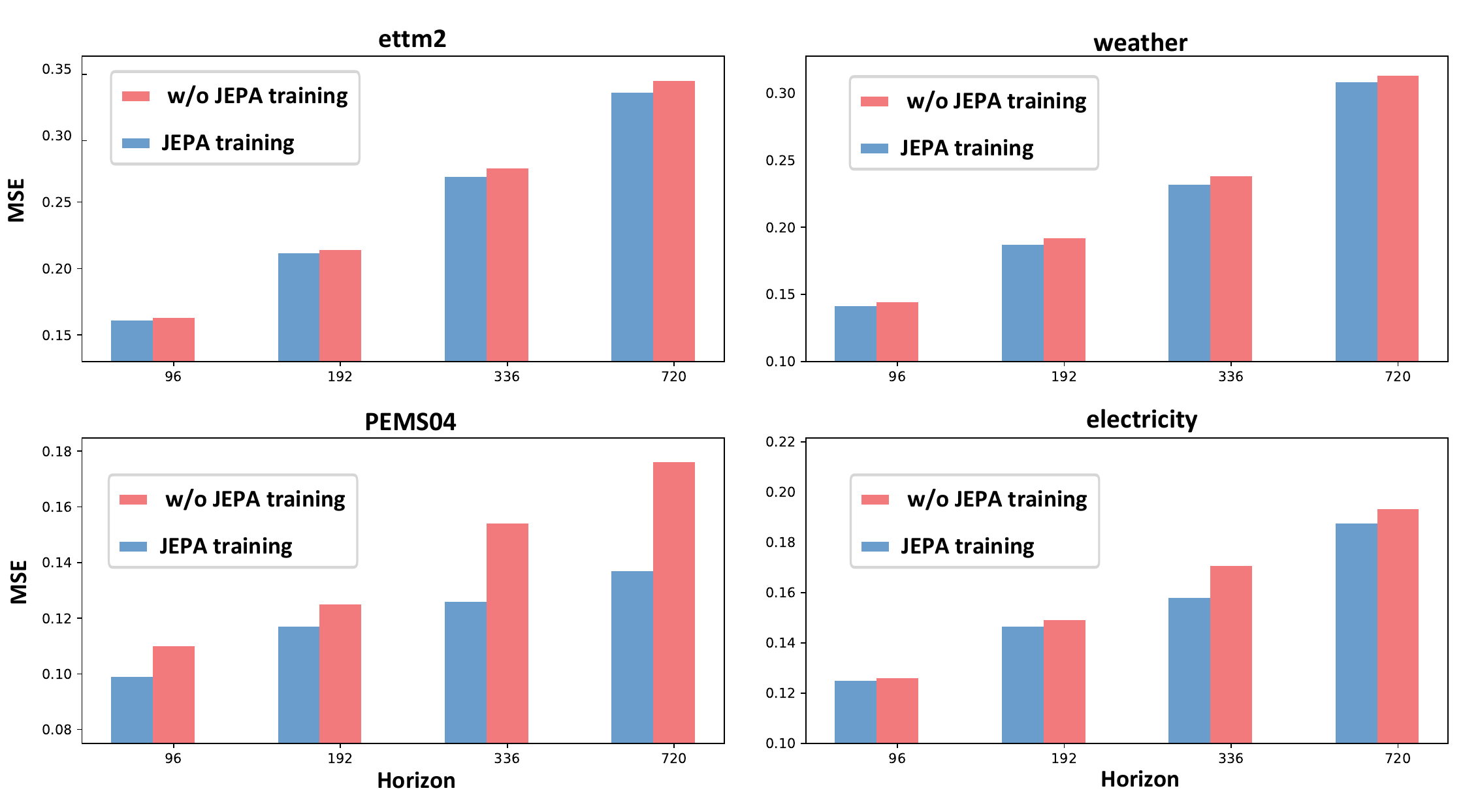}
\end{minipage}
}
\caption{The top figures show the variation trend of JEPA loss without backpropagation, while the bottom figures show the effect of including JEPA loss as a training objective on performance.}
\label{fig:3}
\end{figure}

\begin{figure}[h]
\subfigure[TimeCapsule without JEPA training]{
\begin{minipage}[hbtp]{\linewidth}
\includegraphics[width=\linewidth]{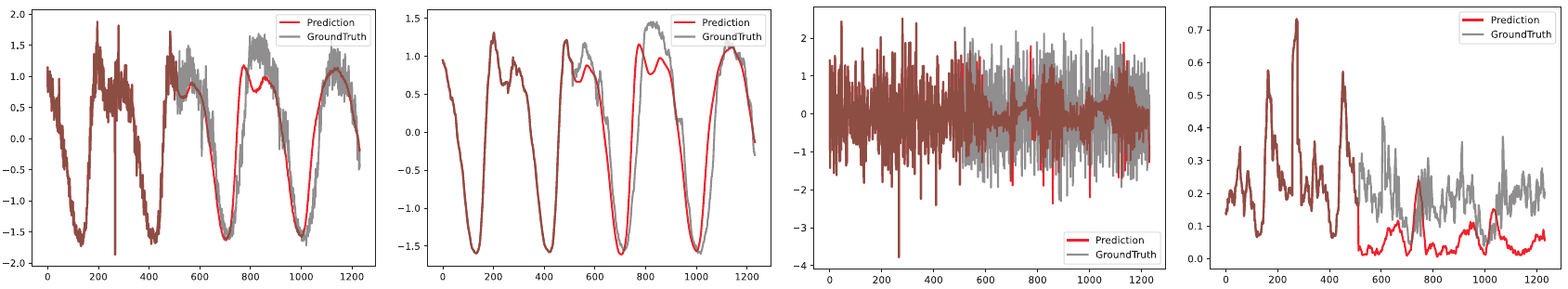}
\end{minipage}
}
\subfigure[TimeCapsule with JEPA training]{
\begin{minipage}[hbtp]{\linewidth}
\includegraphics[width=\linewidth]{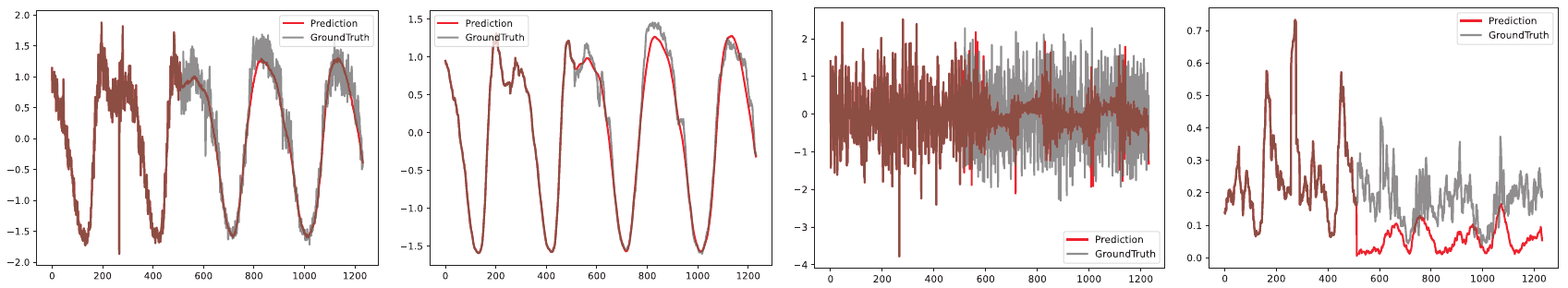}
\end{minipage}
}
\caption{Visualizations of 512-720 prediction on PEMS04 dataset to show the variance of the utilization of JEPA.}
\label{fig:x}
\end{figure}
For the LTSF task, predicting the long-term trend of the time series is paramount, as the data often exhibits complex non-linearity and non-stationarity. To better understand the impact of internal prediction, we approximately decompose the time series into three components: stationary, trend, and deviation, using a simple mean filter for abstraction. As shown in Fig. \ref{fig:y}, the prediction for the trend is indeed crucial and challenging.

We compare TimeCapsule with other deep learning techniques designed to handle non-stationarity in time series. The results show that TimeCapsule outperforms other models in trend prediction. Interestingly, for the ETTh1 dataset, MLP achieves similar performance to TimeCapsule, though its trend predictions are not as sound as TimeCapsule. This difference can be attributed to two factors. First, While not always accurate, TimeCapsule's implicit internal forecasting mechanism allows it to inherently predict the non-stationary trend, alongside other models designed for such tasks. Second, the use of RevIn helps TimeCapsule predict at the correct scales.

Moreover, the JEPA loss function plays a key role in tracking the learning process of predictive representations and can be incorporated into the training to aid in the convergence of internal predictions. However, removing JEPA from backpropagation reveals that its effect is subtle and context-dependent. As illustrated in Fig. \ref{fig:3}, the curve of JEPA loss consistently decreases even without explicit training. This phenomenon partially confirms the existence of a gap between historical compression and future forecasting. The difference in converged JEPA loss between models trained with and without JEPA roughly reflects the distance between two subsets of the metric space. In essence, the optimization direction indicated by JEPA loss minimization is promising.

In contrast, the results for the PEMS04 and Electricity datasets show a growing trend in JEPA loss. To further explore this, we compare forecasting accuracy with their original values at the bottom part of Fig. \ref{fig:3}. From this, we can empirically conclude that in many cases, regularizing the optimization with the inductive bias introduced by JEPA can be beneficial. Incorporating JEPA loss into the training process can encourage the forecaster to converge on a favorable minimum before stepping into other directions. However, the effect of attaching such a reinforcement becomes negligible when the descent directions of two losses align or when a flat energy plane is reached. This finding may provide greater flexibility in the use of JEPA for optimization guidance in predictive representation learning.
\begin{table}[htbp]
\caption{Investigation on compression dimensions. The dimension set contains lengths of compression dimension of \{T(temporal), V(variate), L(level)\}. MSE and MAE are both reported by averaging those from all prediction horizons, and $\swarrow$ denotes performance decline while \textemdash~ means that there are almost no changes in performance.}
\label{tab:4}
\centering
\resizebox{.48\textwidth}{!}{\begin{tabular}{c|cc|cc|cc|cc}
\toprule
\multirow{3}{*}{Dimension Set} & \multicolumn{2}{c}{PEMS04} & \multicolumn{2}{c}{Weather} & \multicolumn{2}{c}{Electricity} & \multicolumn{2}{c}{Traffic}  \\
\cmidrule(r){2-3} \cmidrule(r){4-5} \cmidrule(r){6-7} \cmidrule(r){8-9}  
& MSE~(avg) & MAE~(avg) & MSE~(avg) & MAE~(avg) & MSE~(avg) & MAE~(avg) & MSE~(avg) & MAE~(avg)\\
\midrule
(4, 8, 4) & 0.120 & 0.223 & 0.219 & 0.253 & 0.154 & 0.248 & 0.392 & 0.262 \\
\midrule
(4, 1, 4) & 0.136 & 0.236 & 0.220 & 0.255 & 0.156 & 0.251 & 0.404 & 0.277  \\
\midrule
(1, 1, 1) & 0.156 & 0.257 & 0.220 & 0.254 & 0.158 & 0.250 & 0.410 & 0.285 \\
\midrule
Average Variation (\%) & $\swarrow$ 21.7  & $\swarrow$ 10.6 &\textendash & \textendash & \textendash & \textendash & $\swarrow$ 3.9 & $\swarrow$ 7.2\\
\bottomrule
\end{tabular}}
\end{table}
\subsubsection{Analysis of Asymmetric Structure.} We investigate the asymmetric structure of TimeCapsule by varying the compression dimensions to figure out which part—transformer-based encoder or MLP-based decoder—plays a more crucial role. By default, the compression dimensions are set to \{4, 8, 4\}. Then we test two alternative configurations: (a) \{4, 1, 4\}, which removes the level embedding effect, and (b) \{1, 1, 1\}, which reduces the entire model to a pure MLP. The results in Table \ref{tab:4} reveal that for datasets with higher sampling frequencies, such as PEMS04, multi-level modeling and capturing multi-mode dependencies show significant benefits. In contrast, for datasets like weather and electricity, a simpler MLP structure appears to be sufficient for long-term forecasting. These observations, along with the comprehensive results in Table \ref{tab:1}, suggest that designing a universal, efficient model for diverse datasets is challenging, often leading to inefficient module allocation. Furthermore, a forecaster with strong generalized linear modeling capacity can manage most cases effectively and is more likely to be the key to LTSF. 
\begin{figure*}[th]
\subfigure{
\begin{minipage}[t]{.33\linewidth}
\includegraphics[width=\linewidth]{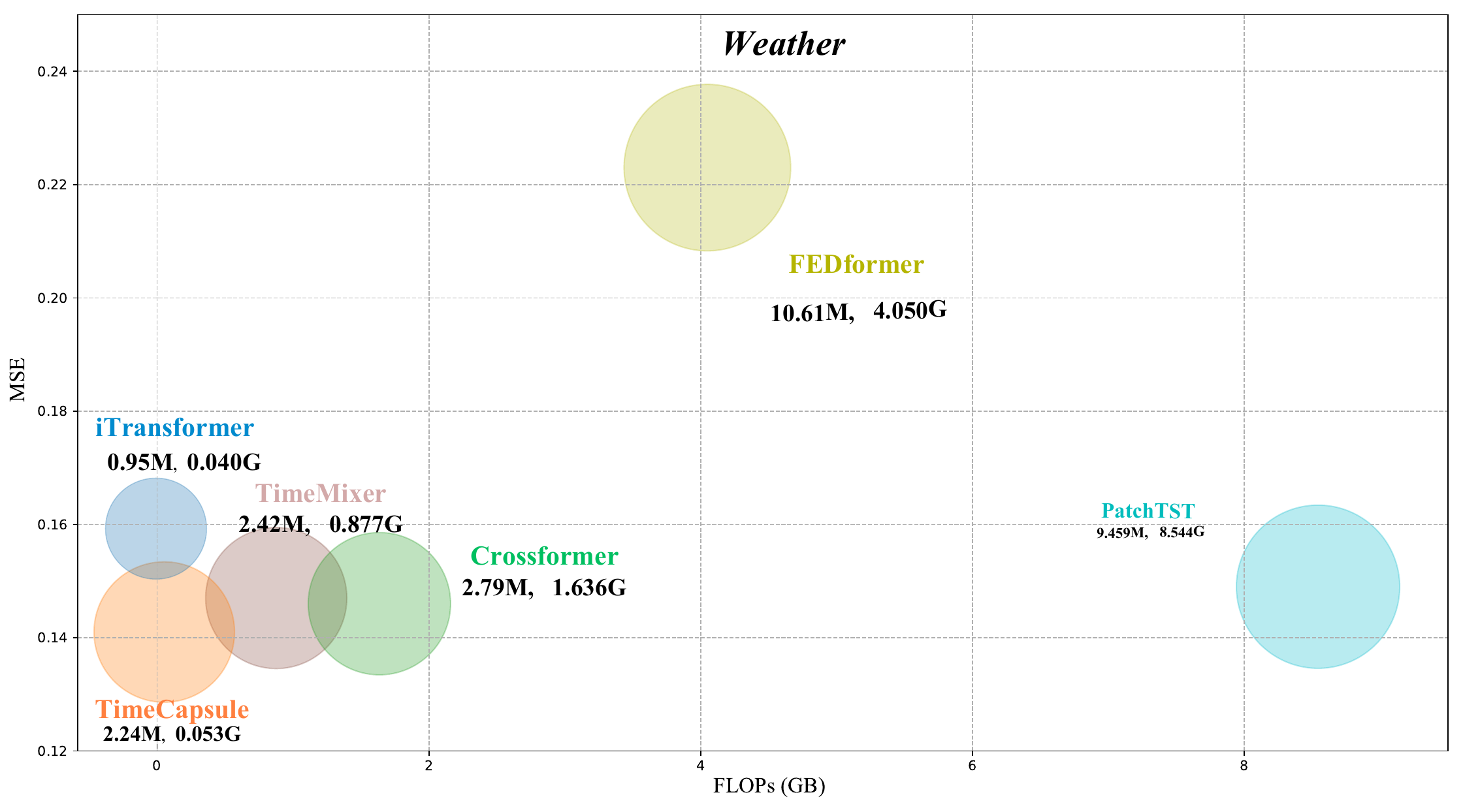}
\end{minipage}
}\subfigure{
\begin{minipage}[t]{.33\linewidth}
\includegraphics[width=\linewidth]{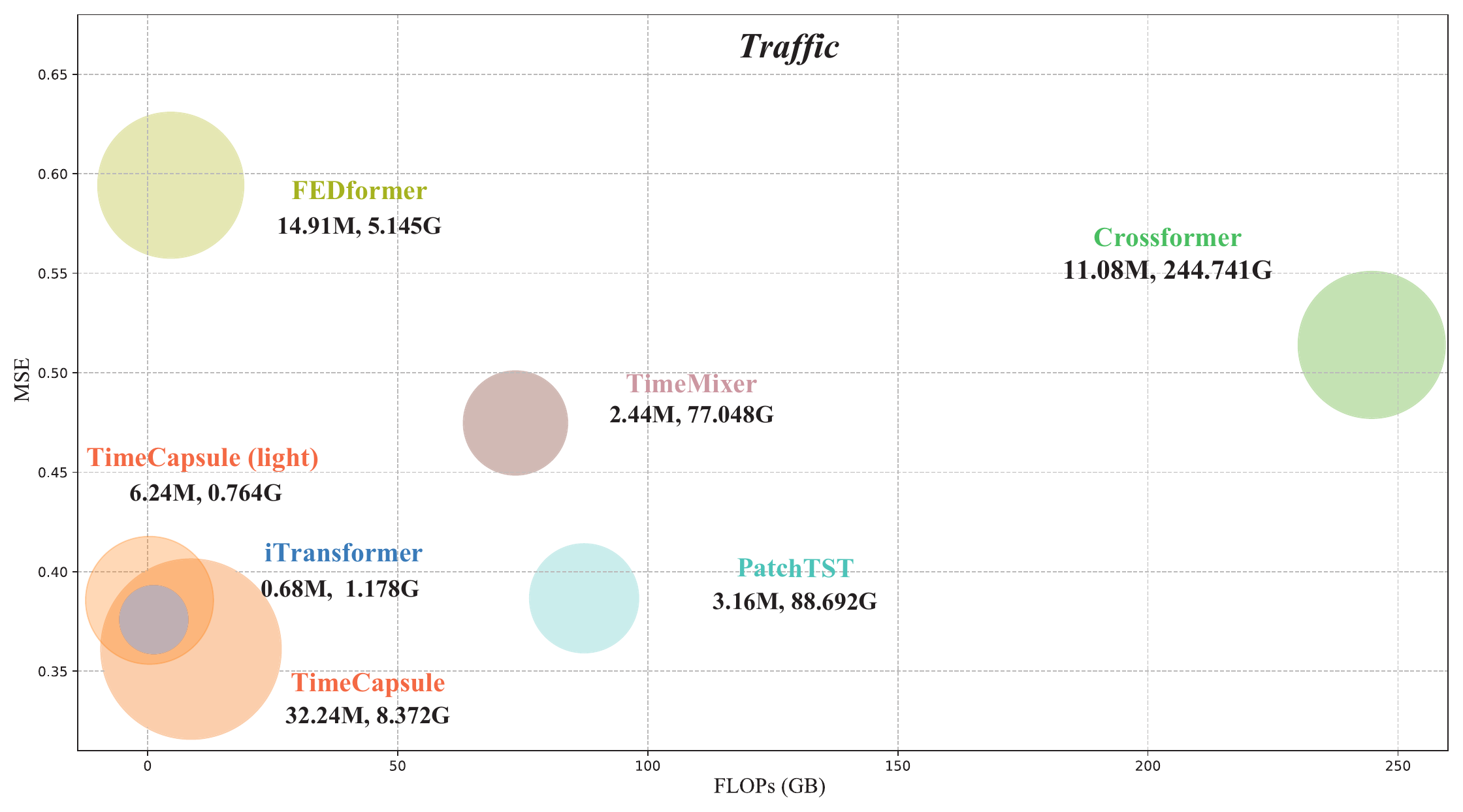}
\end{minipage}
}\subfigure{
\begin{minipage}[t]{.33\linewidth}
\includegraphics[width=\linewidth]{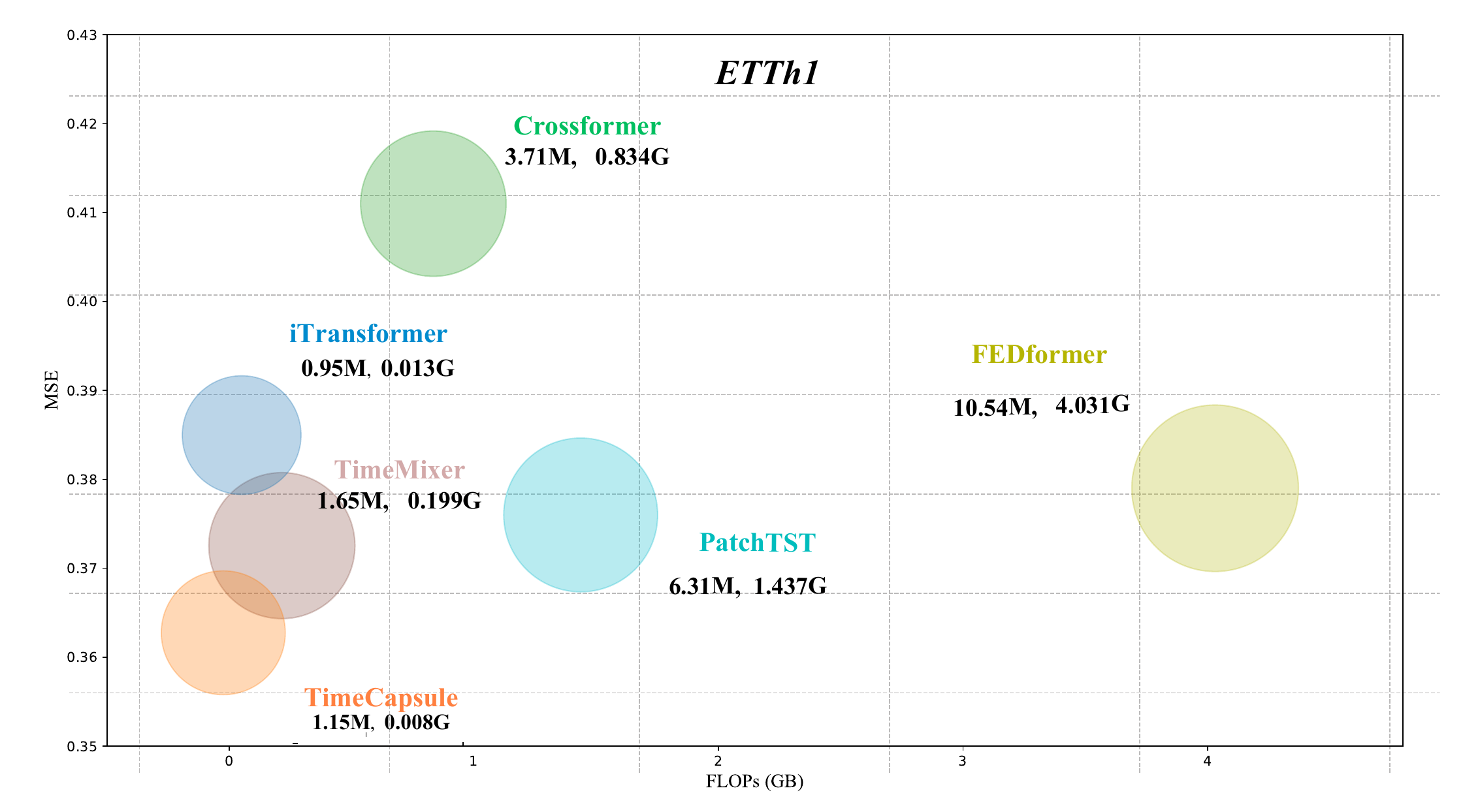}
\end{minipage}
}
\caption{Efficiency comparisons in terms of FLOPs (GB) and parameter counts (MB) with the latest advanced models on the ETTh1, weather, and traffic datasets. Statistics for each model are obtained under the default optimal settings.}
\label{fig:4}
\end{figure*}

\subsection{Efficiency study} Despite TimeCapsule containing multiple transformer blocks in three dimensions, it maintains a reasonable computational complexity due to the use of dimensionality compression and MLPs. To confirm this, we conduct efficiency comparisons on two datasets, Weather and Traffic, analyzing both FLOPs and parameter counts. As shown in Fig.  \ref{fig:4}, our model demonstrates clear advantages in computation speed, though memory usage can sometimes be high. 

Nevertheless, a principal advantage of TimeCapsule is the efficient-performance balance attained by its asymmetric structure, which allows a substantial reduction in parameters without much sacrifice in performance. For instance, we have reduced the hyperparameters, including the number of layers and the number of neurons, for the purpose of forecasting on the Traffic dataset. As shown in the figure, a light version of TimeCapsule attains an MSE of 0.373, compared to the original result of 0.361 (the version incorporating JEPA training). However, in the meantime, both FLOPs and memory usage undergo substantial reduction.
\section{Conclusion}
We propose and empirically evaluate a generic model called TimeCapsule for long-term multivariate time series forecasting. We avoid incorporating explicit designs focused on the core characteristics of time series modeling, but only leverage the learning capacity of generic deep learning modules, complemented by simple strategies such as 3D tensor modeling and multi-mode transforms. By conceptualizing the forecasting process as an information compression task, we integrate JEPA to guide and detect the learning of predictive representations. Although our approach offers promising results, we believe that there are even more effective ways to implement this framework.

\bibliographystyle{ACM-Reference-Format}


\appendix

\onecolumn
\section{Dataset Descriptions}
\label{appen:a}
Ten widely used public datasets are included in our experiment. ETTh1, ETTh2, ETTm1, and ETTm2 \cite{zhou2021informer} represent datasets recording hourly and 15-minute intervals of Electricity Transformer Temperature. The Weather dataset \cite{wu2021autoformer} contains 21 meteorological features sampled every 10 minutes in Germany, while the Electricity dataset \cite{wu2021autoformer} tracks hourly electricity consumption of 321 customers. The ILI dataset \citep{nie2022patch} records weekly patient counts and influenza-like illness (ILI) ratios. The Solar dataset \cite{lai2018solar} captures 10-minute intervals of solar power production from 137 PV plants in 2006. Traffic \cite{wu2021autoformer} dataset records the hourly road occupancy rates from 862 sensors on San Francisco freeways. Additionally, we include the PEMS04 dataset \cite{liu2022scinet}, which contains public traffic network data from California collected in 5-minute intervals, commonly used in spatio-temporal forecasting. Detailed features and settings of these datasets are presented in Table \ref{tab:5}.
\begin{table}[htbp]
\caption{Descriptions of used multivariate time series datasets. The dim column represents the number of variates, and Split column specifies the train-validate-test splitting ratio for each dataset.}
\label{tab:5}
\centering
\resizebox{.8\textwidth}{!}{
\begin{tabular}{l|c|c|c|c|c}
\toprule
Dataset & Dim & Prediction Length & Split & Frequency & domain\\
\midrule
ETTh1, ETTh2 & 7 & {96, 192, 336, 720} & (6, 2, 2) & Hourly & Electricity \\
\midrule
ETTm1, ETTm2 & 7 & {96, 192, 336, 720} & (6, 2, 2) & 15min & Electricity \\
\midrule
Weather & 21 & {96, 192, 336, 720} & (7, 1, 2) & 10min & Environment \\
\midrule
Electricity & 321 & {96, 192, 336, 720} & (7, 1, 2) & Hourly & Electricity \\
\midrule
Traffic & 862 & {96, 192, 336, 720} & (7, 1, 2) & Hourly & Transportation \\
\midrule
Solar & 137 & {96, 192, 336, 720} & (6, 2, 2) & 10min & Energy \\
\midrule
PEMS04 & 307 & {96, 192, 336, 720} & (6, 2, 2) & 5min & Transportation\\
\midrule
ILI & 7 & {24, 36, 48, 60} & (7, 1, 2) & Weakly & Health\\
\bottomrule
\end{tabular}}
\end{table}

\section{Robustness}
In order to assess the robustness of our method, we report the standard deviation across four different random seeds. We select four datasets that show marginal improvements in forecasting accuracy compared to the second-best method. The results, presented in Table \ref{tab:6}, confirm the reliability of the performance outcomes listed in Table \ref{tab:1}.
\begin{table}[htbp]
  \centering
  \caption{Robustness of TimeCapsule performance. The results are obtained from four random seeds.}
  \label{tab:6}
  \resizebox{\textwidth}{!}{%
    \begin{tabular}{c|cc|cc|cc|cc}
      \toprule
      Dataset & \multicolumn{2}{c|}{traffic} & \multicolumn{2}{c|}{weather} & \multicolumn{2}{c|}{ETTm2} & \multicolumn{2}{c}{electricity}                                                                         \\
      \cmidrule(lr){2-3} \cmidrule(lr){4-5} \cmidrule(lr){6-7} \cmidrule(lr){8-9}
      Horizon & MSE & MAE & MSE  & MAE  & MSE & MAE & MSE & MAE  \\
      \midrule
      96 & 0.363$\pm$0.002  & 0.246$\pm$0.001 & 0.143 $\pm$0.001            & 0.189$\pm$0.002                 & 0.162$\pm$0.001 & 0.250$\pm$0.000 & 0.126$\pm$0.001 & 0.218$\pm$0.000 \\
      192     & 0.383$\pm$0.000              & 0.257$\pm$0.000              & 0.189$\pm$0.003            & 0.233$\pm$0.002                 & 0.218$\pm$0.001 & 0.289$\pm$0.002 & 0.146$\pm$0.001 & 0.238$\pm$0.001 \\
      336     & 0.394$\pm$0.004              & 0.264$\pm$0.006              & 0.241$\pm$0.002            & 0.274$\pm$0.002                 & 0.270$\pm$0.001 & 0.324$\pm$0.001 & 0.162$\pm$0.001 & 0.255$\pm$0.001 \\
      720     & 0.430$\pm$0.000              & 0.282$\pm$0.001              & 0.310$\pm$0.002            & 0.326$\pm$0.002                 & 0.347$\pm$0.002 & 0.376$\pm$0.002 & 0.195$\pm$0.001 & 0.285$\pm$0.001 \\
      \bottomrule
    \end{tabular}%
  }
\end{table}
\section{More Studies}
\subsection{Lookback Window} 
\begin{figure}[htbp]
\centering
  \includegraphics[width=\linewidth]{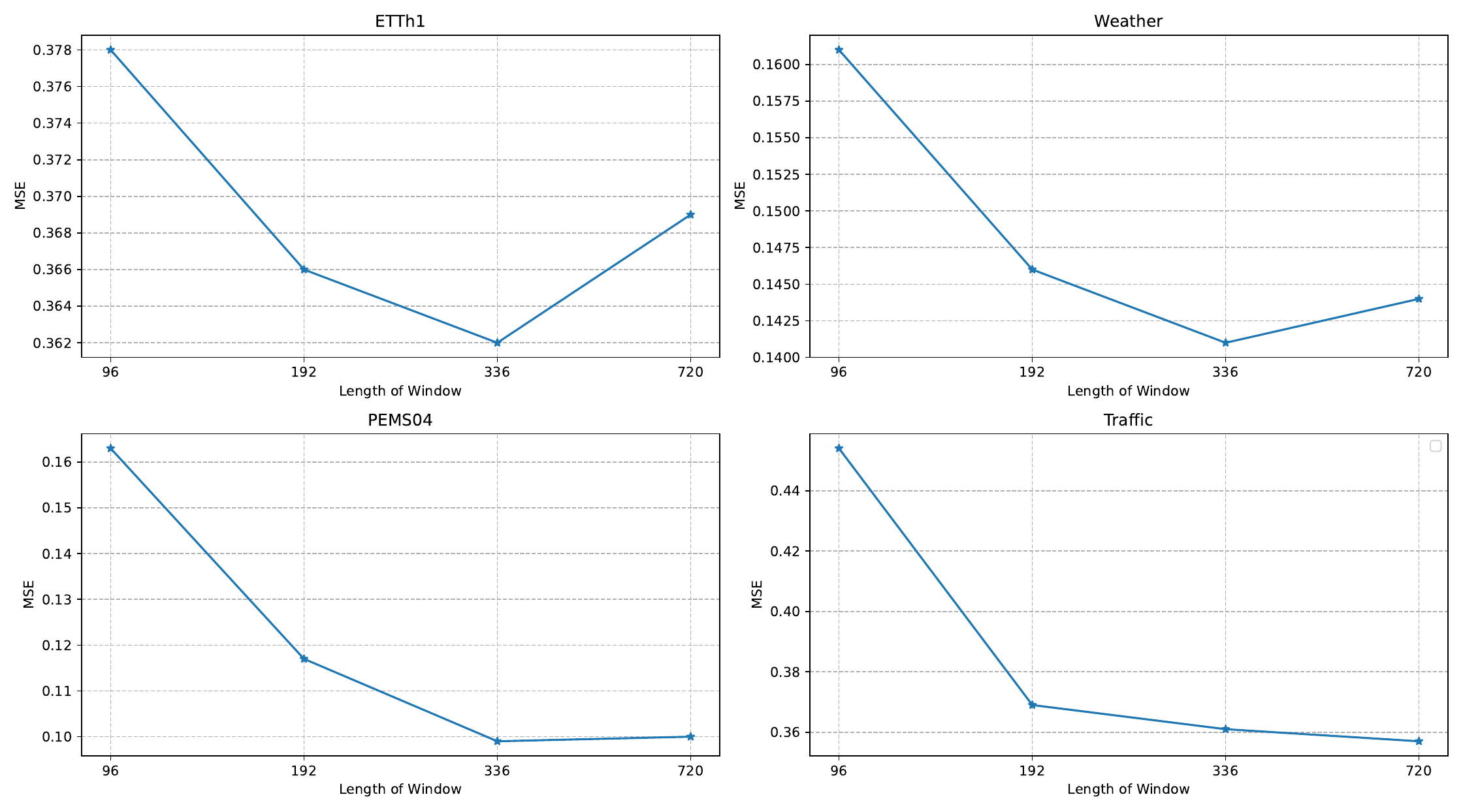}
  \caption{Study on varying lookback windows. We set the length of window $T_x\in\{96, 192, 336, 720\}$, and fix the forecasting horizon $T_y=96$. }
  \label{fig:5}
\end{figure}

This study examines the impact of varying the lookback window on forecasting performance. As illustrated in Fig. \ref{fig:5}, the accuracy consistently improves with an enlarged lookback window, ranging from 96 to 512. However, this effect diminishes for small and medium datasets as the window length increases.
\subsection{Noise in Encoding Tunnels}
To enhance the robustness of the encoding process, we introduced Gaussian noise before the transformation step, as illustrated in Fig. \ref{fig:2} and detailed in formula (\ref{momsa}). To evaluate the impact of this added noise, we conducted an ablation study, which is discussed below.
\begin{table}[hbtp]
\caption{Ablations on the noise added in the encoder. The results are obtained by averaging from all four prediction horizons.}
\label{tab:9}
\centering
\resizebox{\textwidth}{!}{\begin{tabular}{c|cc|cc|cc|cc}
\toprule
\multirow{3}{*}{Method} & \multicolumn{2}{c}{ETTm2} & \multicolumn{2}{c}{Weather} & \multicolumn{2}{c}{Traffic} & \multicolumn{2}{c}{Electricity}\\
\cmidrule(r){2-3} \cmidrule(r){4-5} \cmidrule(r){6-7} \cmidrule(r){8-9}
& MSE (avg) & MAE (avg) & MSE (avg) & MAE (avg) & MSE (avg) & MAE (avg) & MSE (avg) & MAE (avg) \\
\midrule
Origin & 0.247 & 0.308 & 0.219 & 0.253 & 0.392 & 0.262 & 0.157 & 0.249\\
\midrule
w/o noise & 0.250 & 0.310 & 0.224 & 0.258 & 0.392 & 0.262 & 0.219 & 0.265 \\
\bottomrule
\end{tabular}}
\end{table}

As shown in Table \ref{tab:9}, introducing noise generally enhances forecasting performance across the studied datasets. However, the impact varies significantly. For datasets such as ETTm2 and Weather, the improvements are marginal. In contrast, for a large-scale dataset like Traffic, the forecasting results remain unchanged. Notably, for datasets such as Electricity, the improvements are substantial.
\subsection{Inherent order of TransBlocks within the Encoder} We have illustrated the rationale behind the order of TransBlocks in the encoder in Section \ref{feedforward}. To provide further evidence of its reasonability, we undertake an empirical comparison of the results obtained by TimeCapsule with different order settings. Furthermore, in order to maintain the advantage of efficiency, we only demonstrate the performance variations by exchanging the order of L-Block and V-Block. The results on three large datasets are recorded in the following table, which demonstrates that TimeCapsule with the current order exhibits a slight superiority over TimeCapsule with an exchanged block order. This phenomenon serves to validate our assertions regarding the advantage of learning multi-level properties in advance.

\begin{table}[hbtp]
\caption{Ablations on the block order within the encoder. The results are obtained by averaging from all four prediction horizons.}
\centering
\resizebox{\textwidth}{!}{\begin{tabular}{c|cc|cc|cc}
\toprule
\multirow{3}{*}{Method} & \multicolumn{2}{c}{Weather} & \multicolumn{2}{c}{Traffic} & \multicolumn{2}{c}{Electricity}\\
\cmidrule(r){2-3} \cmidrule(r){4-5} \cmidrule(r){6-7} 
& MSE (avg) & MAE (avg) & MSE (avg) & MAE (avg) & MSE (avg) & MAE (avg) \\
\midrule
Origin & 0.219 & 0.253 & 0.392 & 0.262 & 0.157 & 0.249\\
\midrule
Exchanged V-L order & 0.223 & 0.261 & 0.407 & 0.274 & 0.165 & 0.260 \\
\bottomrule
\end{tabular}}
\end{table}

\subsection{Positional Encoding}
\begin{table}[hbtp]
\caption{Ablations on positional encoding, where w/o PE denotes positional encoding. Performance values are averaged from all four forecasting horizons. The results show that forecasting performances decrease when the positional encoding is removed. }
\label{tab:7}
\centering
\resizebox{\textwidth}{!}{\begin{tabular}{c|cc|cc|cc|cc}
\toprule
\multirow{3}{*}{Method} & \multicolumn{2}{c}{ETTm2} & \multicolumn{2}{c}{Weather} & \multicolumn{2}{c}{ETTh2} & \multicolumn{2}{c}{ETTm1}\\
\cmidrule(r){2-3} \cmidrule(r){4-5} \cmidrule(r){6-7} \cmidrule(r){8-9}
& MSE (avg) & MAE (avg) & MSE (avg) & MAE (avg) & MSE (avg) & MAE (avg) & MSE (avg) & MAE (avg) \\
\midrule
Origin & 0.247 & 0.308 & 0.219 & 0.253 & 0.338 & 0.386 & 0.345 & 0.376\\
\midrule
w/o PE & 0.252 & 0.312 & 0.222 & 0.255 & 0.346 & 0.391 & 0.349 & 0.379 \\
\bottomrule
\end{tabular}}
\end{table}
In our default settings, temporal positional encoding is applied at the head of the T-TransBlock. We question whether attaching this auxiliary information remains beneficial in the compressed representation space, as its effect could also be an obscure signal to evidence the efficacy of our information compression mechanism. The ablation results, presented in Table \ref{tab:7}, show that excluding positional encoding will lead to a slight decline in performance. This suggests that although the impact is minor due to the reduced dimensionality, positional encoding still contributes valuable temporal information even in the compressed representation space. 

We can speculate that the effectiveness of positional encoding stems from the linear nature of our compression, which is simply implemented through multiplication by a low-rank transform matrix. This can be illustrated by the following toy example:
\begin{equation*}
    \textbf{M}(\text{X}+\text{PE})=\textbf{M}(\text{X})+\textbf{M}(\text{PE})
\end{equation*}
where $\textbf{M}$ represents the linear operator, and $\text{X}$ and $\text{PE}$ correspond to the input data and additive positional encoding, respectively. This demonstrates that the positional encoding is projected into the same compressed transformation space as the input data.
\subsection{Compression Dimension}
\label{appen:d3}
\begin{table}[htbp]
\caption{Full results of the study on compression dimensions. The dimension set contains lengths of compression dimension of \{T(temporal), V(variate), L(level)\}. MSE and MAE are both reported by averaging those from all prediction horizons, and $\swarrow$ denotes performance decline while \textemdash~ means no significant changes in performance.}
\label{tab:8}
\centering
\resizebox{\textwidth}{!}{\begin{tabular}{c|cc|cc|cc|c}
\toprule
\multirow{1}{*}{Dimension Set} & \multicolumn{2}{c}{origin} & \multicolumn{2}{c}{(-, 1, -)} & \multicolumn{2}{c|}{(1, 1, 1)} & \multirow{3}{*}{Trend}  \\
\cmidrule(r){2-3} \cmidrule(r){4-5} \cmidrule(r){6-7}
Metric & MSE~(avg) & MAE~(avg) & MSE~(avg) & MAE~(avg) & MSE~(avg) & MAE~(avg)\\
\midrule
PEMS04 & 0.120 & 0.223 & 0.136 & 0.236 & 0.156 & 0.257 & $\swarrow$ \\
\midrule
Traffic & 0.392 & 0.262 & 0.404 & 0.277 & 0.410 & 0.285 & $\swarrow$ \\
\midrule
Weather & 0.219 & 0.253 & 0.220 & 0.255 & 0.220 & 0.254 & \textemdash \\
\midrule
Electricity & 0.157 & 0.249 & 0.158 & 0.251 & 0.159 & 0.250 & \textemdash \\
\midrule
Solar & 0.191 & 0.243 & 0.193 & 0.244 & 0.191 & 0.242 & \textemdash \\
\midrule
ETTh1 & 0.408 & 0.428 & 0.413 & 0.430 & 0.419 & 0.433 & $\swarrow$  \\
\midrule
ETTm1 & 0.345 & 0.376 & 0.348 & 0.377 & 0.348 & 0.378 & \textemdash \\
\midrule
ETTh2 & 0.339 & 0.387 & 0.356 & 0.396 & 0.357 & 0.398 & $\swarrow$  \\
\midrule
ETTm2 & 0.248 & 0.309 & 0.251 & 0.311 & 0.250 & 0.311 & \textemdash  \\
\bottomrule
\end{tabular}}
\end{table}
We provide additional experimental results exploring the Transformer-MLP trade-off by varying the compression dimensions. As shown in Table \ref{tab:8}, for most datasets, a robust MLP architecture without explicit dependency capturing is sufficient to achieve relatively strong performance. This supports our argument that forecasting models often suffer from inefficient resource allocation when processing diverse datasets. A flexible structure, such as that of TimeCapsule, proves effective in adapting to different scenarios.
\newpage
\section{What have learnt by TimeCapsule ?}
\label{learn}
One of the most exciting aspects of TimeCapsule is its ability to handle time series decomposition autonomously through neural networks, potentially at the expense of interpretability. In this part, we explore the decomposition strategy employed by TimeCapsule through visualizations, aiming to both clarify the inner workings of our model and inspire further investigations into time series modeling with deep learning.

We choose the ETTm2 and PEMS04 datasets to represent, respectively, simpler and more complex time series patterns. Our investigation centers on addressing three key questions:

1. Do the transforms retain and differentiate variable and level information?
 
2. What do the transform matrices reveal?

3. What kind of decomposition strategies has TimeCapsule learnt?
 
\subsection{Do these transforms retain and distinguish the information of variables and levels ?}
\begin{figure}[htbp]
\centering
  \includegraphics[width=\linewidth]{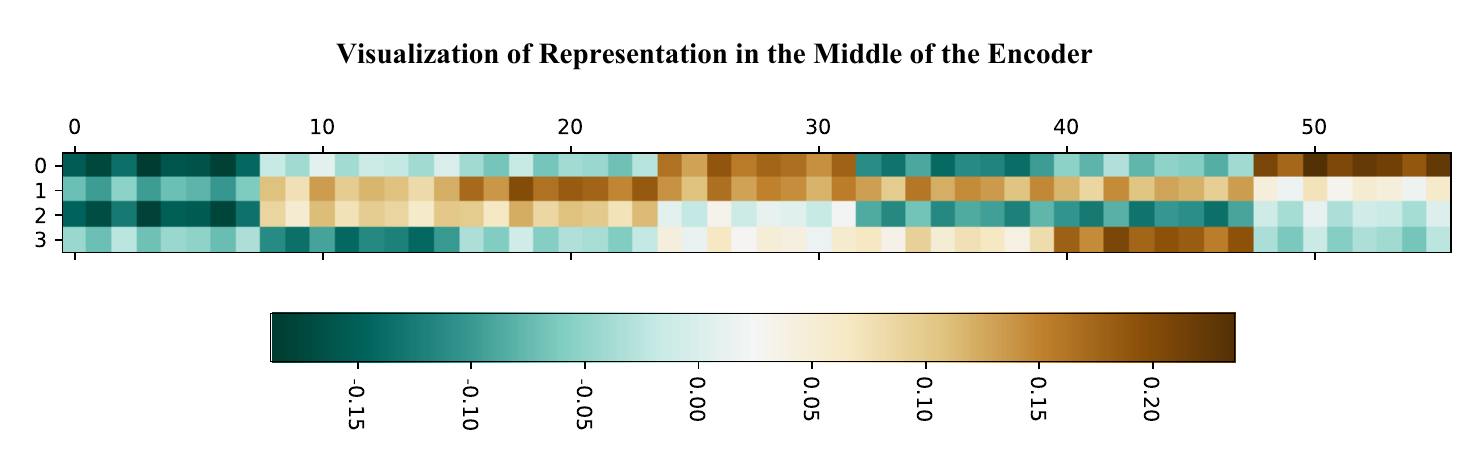}
  \caption{Representaion of ETTm2 in the encoder. We select $\mathcal{X}_2 \in \mathbb{R}^{4 \times 8\times 7}$ at the end of the $\text{L-TransBlock}$. It has the compressed temporal dimension $t_c = 4$, expanded level dimension $l_c=8$, and variable dimension $v=7$. }
  \label{fig:6}
\end{figure}
\begin{figure}[htbp]
\centering
  \includegraphics[width=\linewidth]{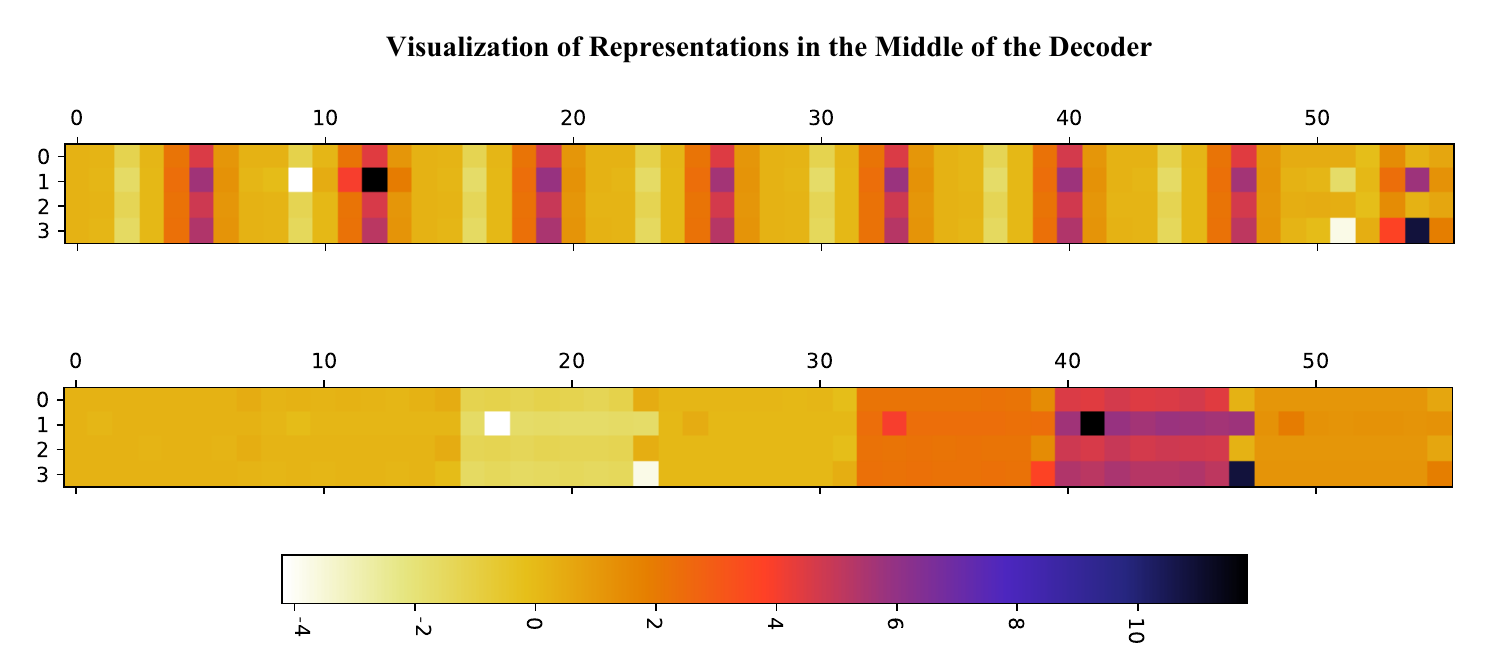}
  \caption{Representaion of ETTm2 in the decoder. We select $\mathcal{Y}_1 \in \mathbb{R}^{4 \times 8\times 7}$, which also has the compressed temporal dimension $t_c = 4$, expanded level dimension $l_c=8$, and variable dimension $v=7$. }
  \label{fig:7}
\end{figure}
Firstly, as shown in Fig. \ref{fig:6}, we present the representation obtained after the time compression and level expansion, denoted as $\mathcal{X}_2\in \mathbb{R}^{v\times T_c\times l}$ with $v=4$, $t_c=$,4 and $l=8$. By folding it into the matrix  $\mathbf{X}\in \mathbb{R}^{t_c\times lv}$, we observe that even in the compressed representation space, distinct characteristics of variables remain. These are organized into seven recognized blocks, each may correspond to a variable in the ETTm2 dataset. What's more, each block contains eight components, which can be interpreted as level tokens.

In the decoder stage, we examine the representation $\mathcal{Y}_1\in \mathbb{R}^{v\times T_c\times l}$ in the same way. As illustrated in the bottom half of Fig. \ref{fig:7}, though patterns differ from those in $\mathcal{X}_2$, there are still seven blocks. 
Furthermore, when we swap the dimension of $l$ and $T_c$ in $\mathcal{Y}_1$ (top part of Fig. \ref{fig:7}), the representation turns out to have eight blocks with seven components within each, which exactly align with our layout depicted in the leftmost part of Fig. \ref{fig:2}. This finding is intriguing: despite the compression of the variable dimension, and without explicitly instructing TimeCapsule to learn level and variable tokens separately, it inherently does so.   This suggests that the representation is learned in a predictive and structured manner, indicating that the mode-specific self-attention mechanism is functioning and that multi-level dependencies are effectively captured.

Besides, Fig. \ref{fig:7} also reveals that most levels appear redundant, explaining the results in Table. \ref{tab:8},  where multi-level modeling shows minimal benefit for ETTm2.
\begin{figure}[htbp]
\subfigure{
\begin{minipage}[tb]{.5\linewidth}
\includegraphics[width=\linewidth]{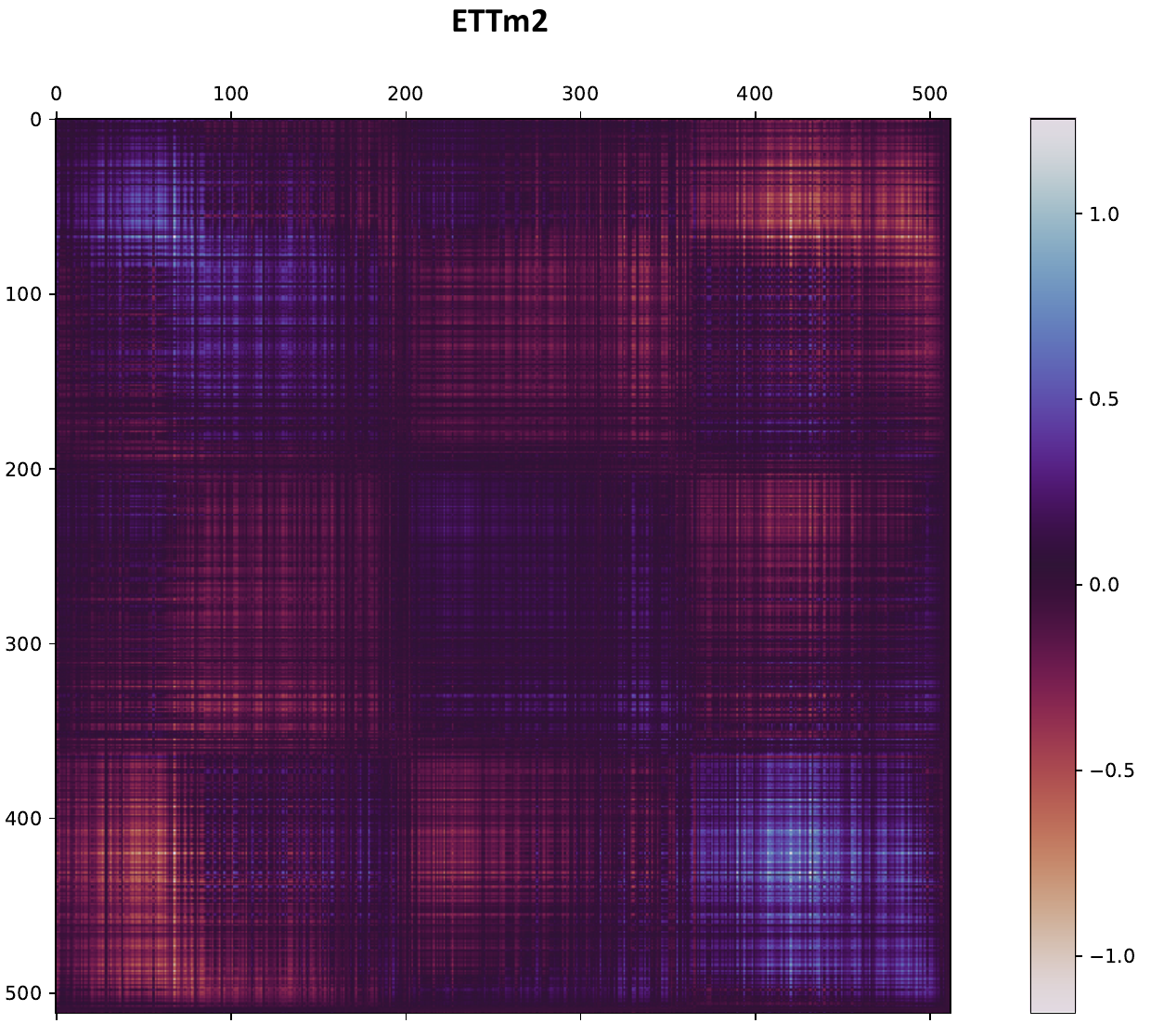}
\end{minipage}
}\subfigure{
\begin{minipage}[tb]{.5\linewidth}
\includegraphics[width=\linewidth]{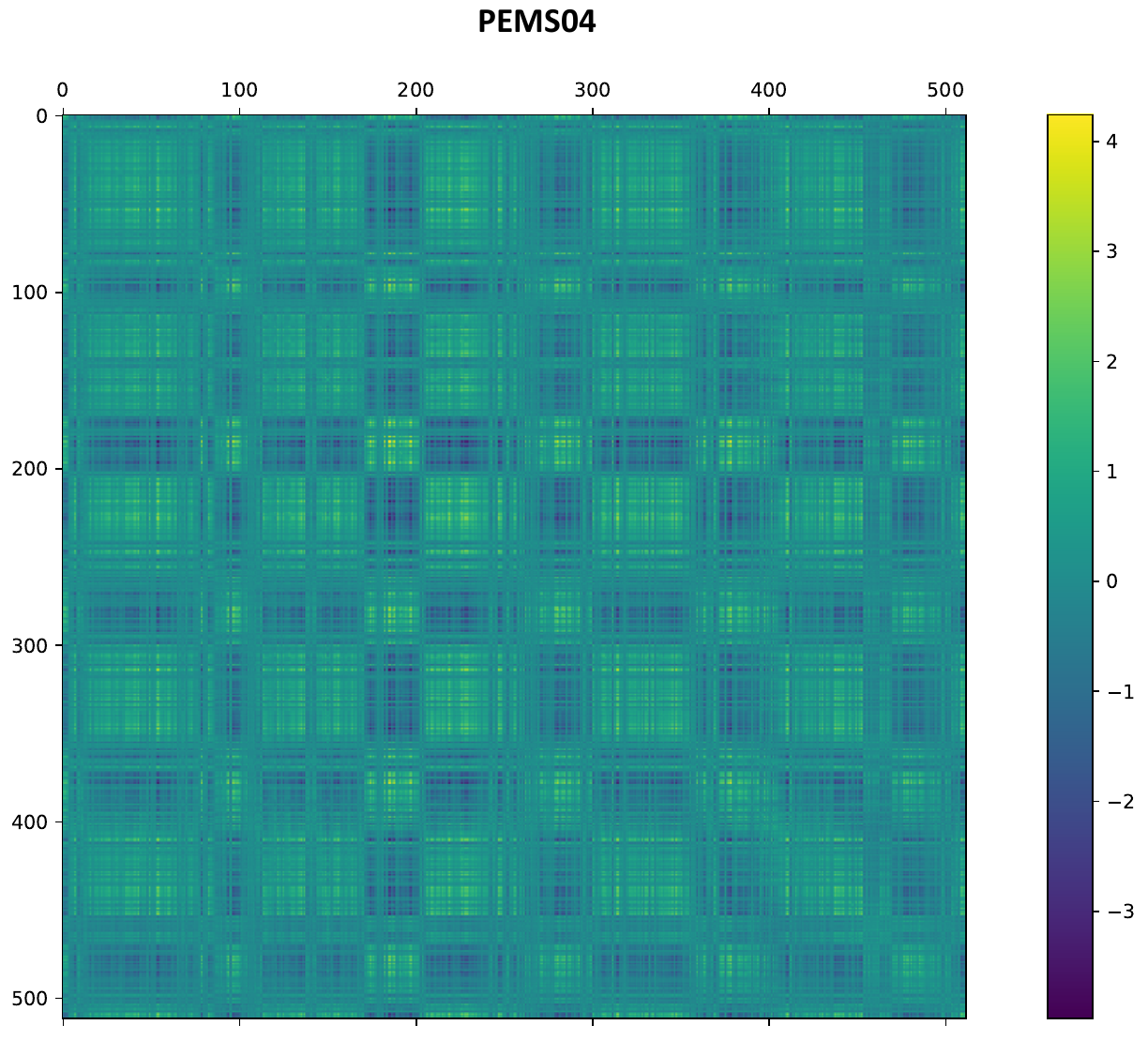}
\end{minipage}
}
\caption{Visualization of T (temporal)-transform matrices, i.e., $\mathbf{M}_\text{T}\mathbf{M}_\text{T}^\top$. The left part shows the learned temporal pattern of ETTm2 dataset, while the right part shows that of PEMS04 dataset.}
\label{fig:8}
\vskip -0.2in
\end{figure}
\begin{figure}[htbp]
\subfigure{
\begin{minipage}[tb]{.5\linewidth}
\includegraphics[width=\linewidth]{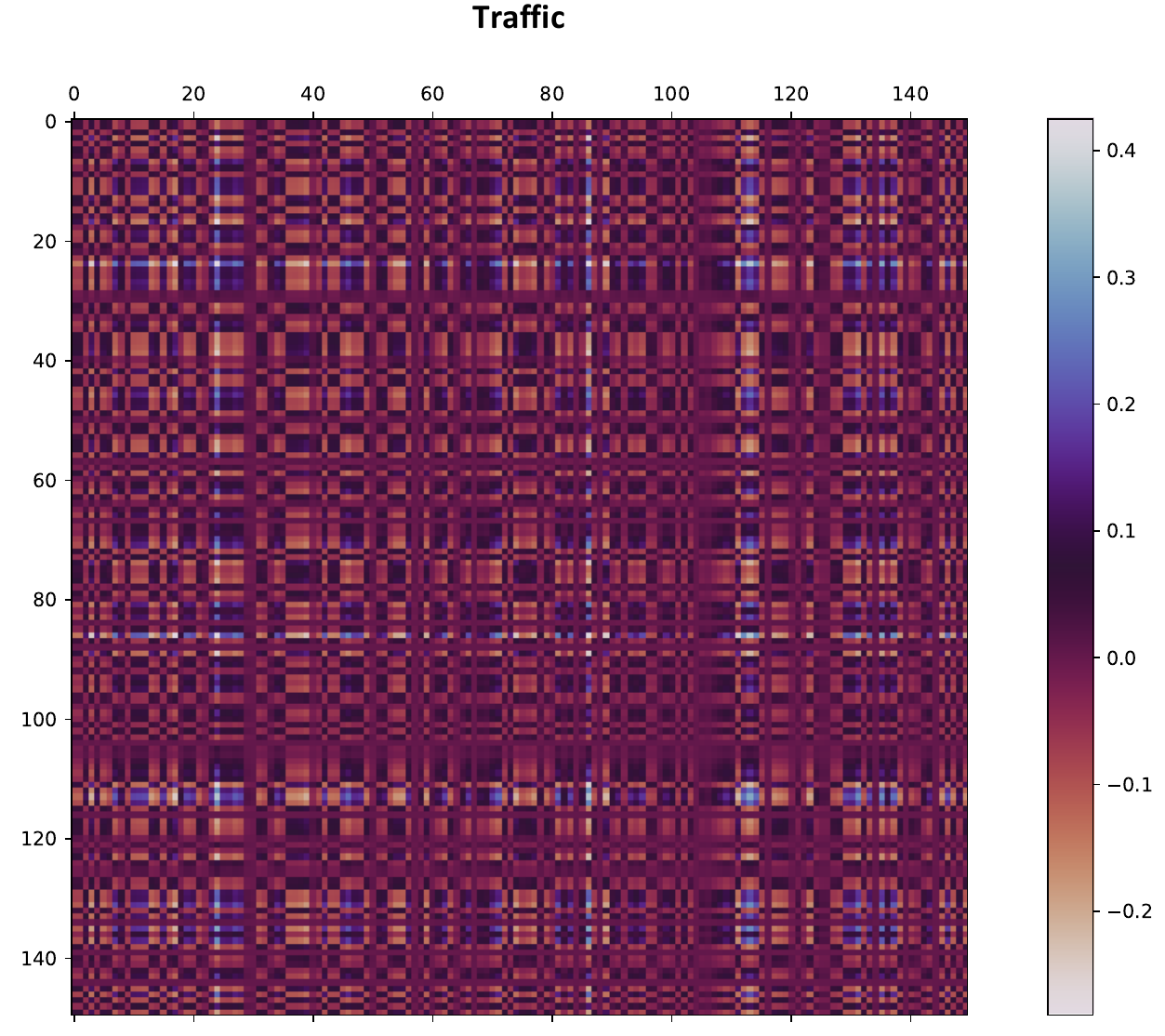}
\end{minipage}
}\subfigure{
\begin{minipage}[tb]{.5\linewidth}
\includegraphics[width=\linewidth]{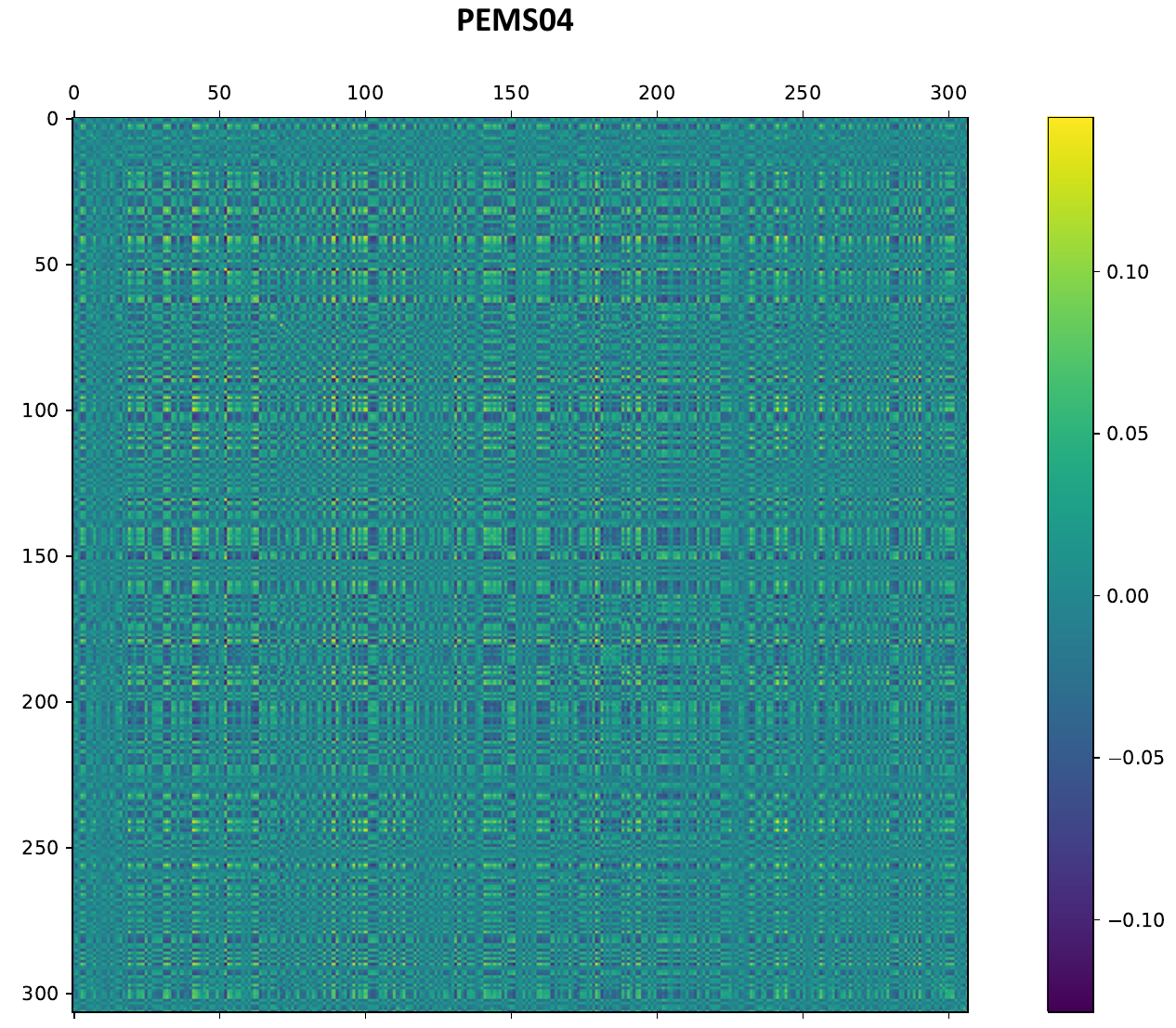}
\end{minipage}
}
\caption{Visualization of V (variable)-transform matrices, i.e., $\mathbf{M}_\text{V}\mathbf{M}_\text{V}^\top$. Since the number of variables contained in ETTm2 are too small to demonstrate a significant pattern, we replace it with Traffic dataset.}
\label{fig:9}
\end{figure}
\subsection{What do these transform matrices look like to enable multi-mode and multi-level learning ?}
The capacity for information compression reflects the model's ability to capture and aggregate dependencies.  As expected, these matrices should exhibit a certain block structure to capture independent and hierarchical features. For clarity, we denote the transform matrices as $\mathbf{M}$ and present them in symmetric form as $\mathbf{MM}^\top$. We omit the level expansion transformation matrix, as its $\mathbf{MM}^\top$ results in a scalar value.

As illustrated clearly in Fig. \ref{fig:8} and Fig. \ref{fig:9}, these low-rank transforms exhibit significant patterns, demonstrating our model's strategies to capturing temporal and variable dependencies within time series.
These visualizations of learned compression/expansion matrices hold promising potential for analyzing the unique characteristics of different time series.

\subsection{What decomposition strategies has TimeCapsule learned ?}
\begin{figure}[htbp]
\centering
  \includegraphics[width=\linewidth]{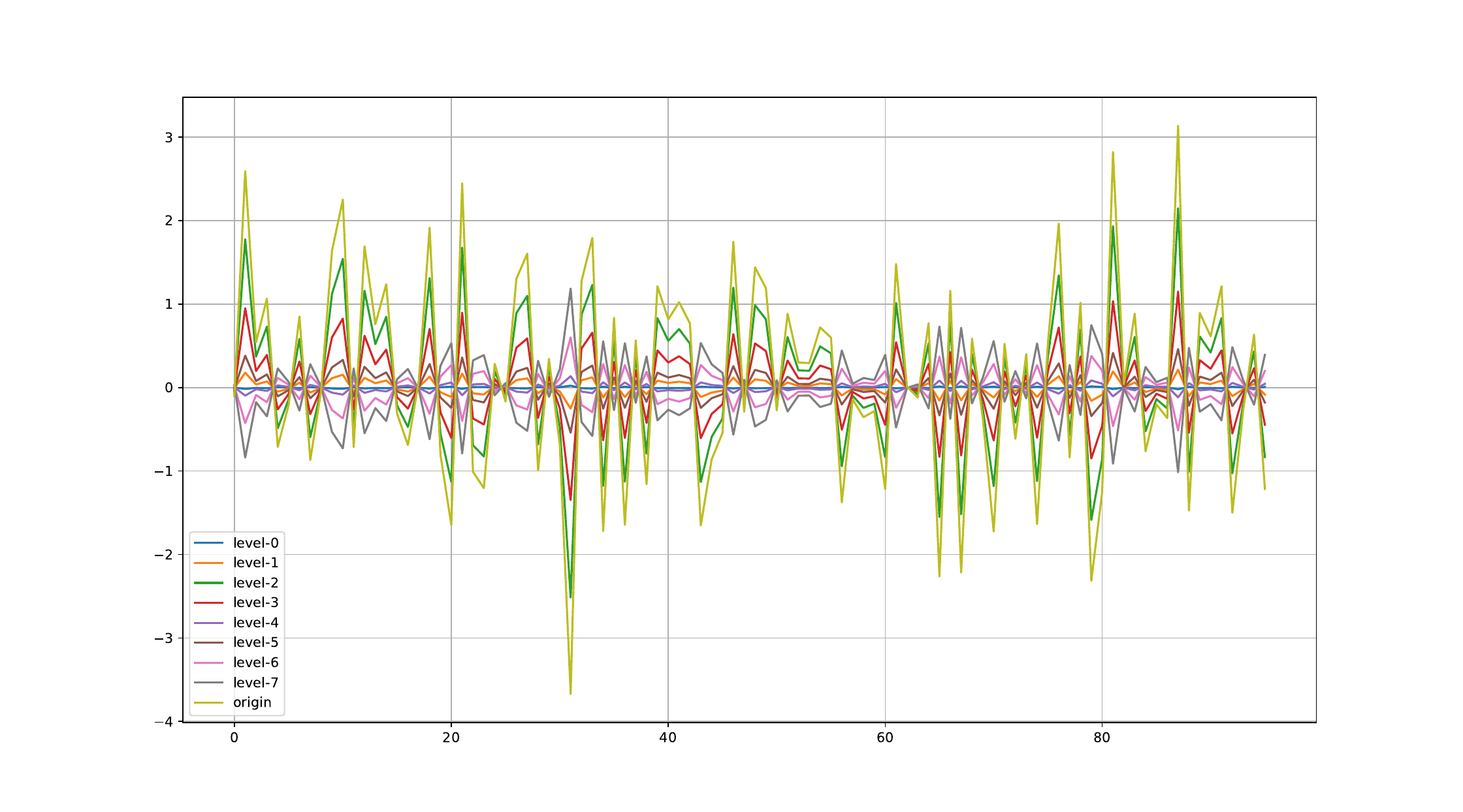}
  \caption{Multil-level property of PEMS04 series in the representation space by applying the learnd level-expansion matrix.}
  \label{fig:11}
\end{figure}

\begin{figure}[htbp]
\centering
  \includegraphics[width=\linewidth]{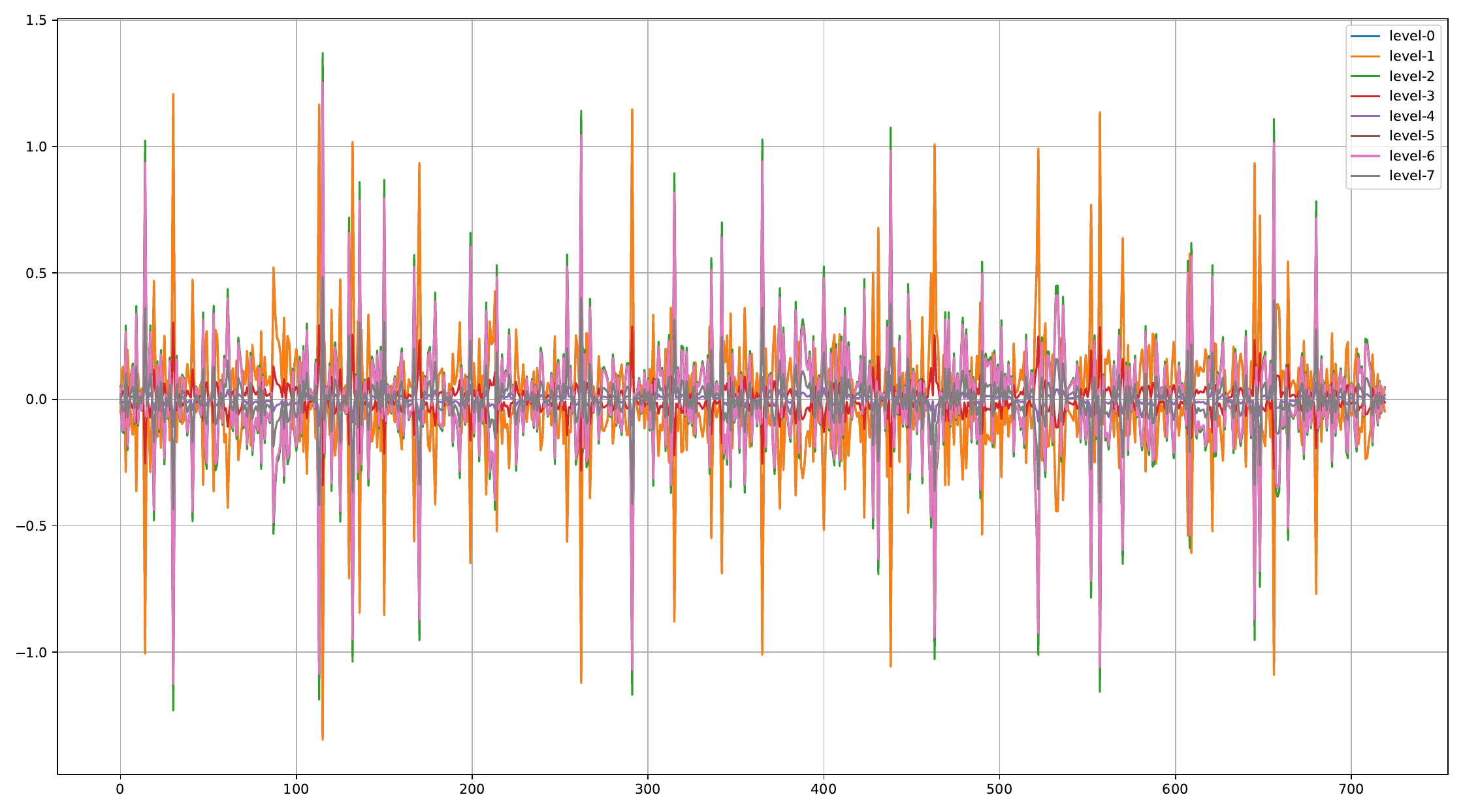}
  \caption{Multil-level property of ETTm2 series in the representation space by applying the learnd level-expansion matrix.}
  \label{fig:10}
\end{figure}

We have observed that TimeCapsule recognizes the multi-level structure of time series, although this property resides in the latent representation space. In order to gain insight into it, we visualize the final representation $\mathcal{Y}_3\in \mathbb{R}^{v\times t_y\times 1}$ at the neck of the decoder. By applying the learned level expansion transform denoted as $\mathbf{M}_\text{L}\in\mathbb{R}^{1\times l}$, we decompose $\mathcal{Y}_3$ into $l$ sub-level series by performing the mode-3 product $\mathcal{Y}_3\times_3 \mathbf{M}_\text{L}$, then display each one, fixing on the first variable. As seen in Fig. \ref{fig:10} and Fig. \ref{fig:11},  these series generally exhibit different scales, indicating that the multi-level property contains a multi-scale property within the representation space. Specifically, for the ETTm2 dataset (see Fig. \ref{fig:10}), each level's series has a unique amplitude, and their frequencies group into different ranges; whereas for PEMS04 (see Fig. \ref{fig:11}), the level patterns appear clearer and simpler.
\begin{figure}[htbp]
\centering
  \includegraphics[width=\linewidth]{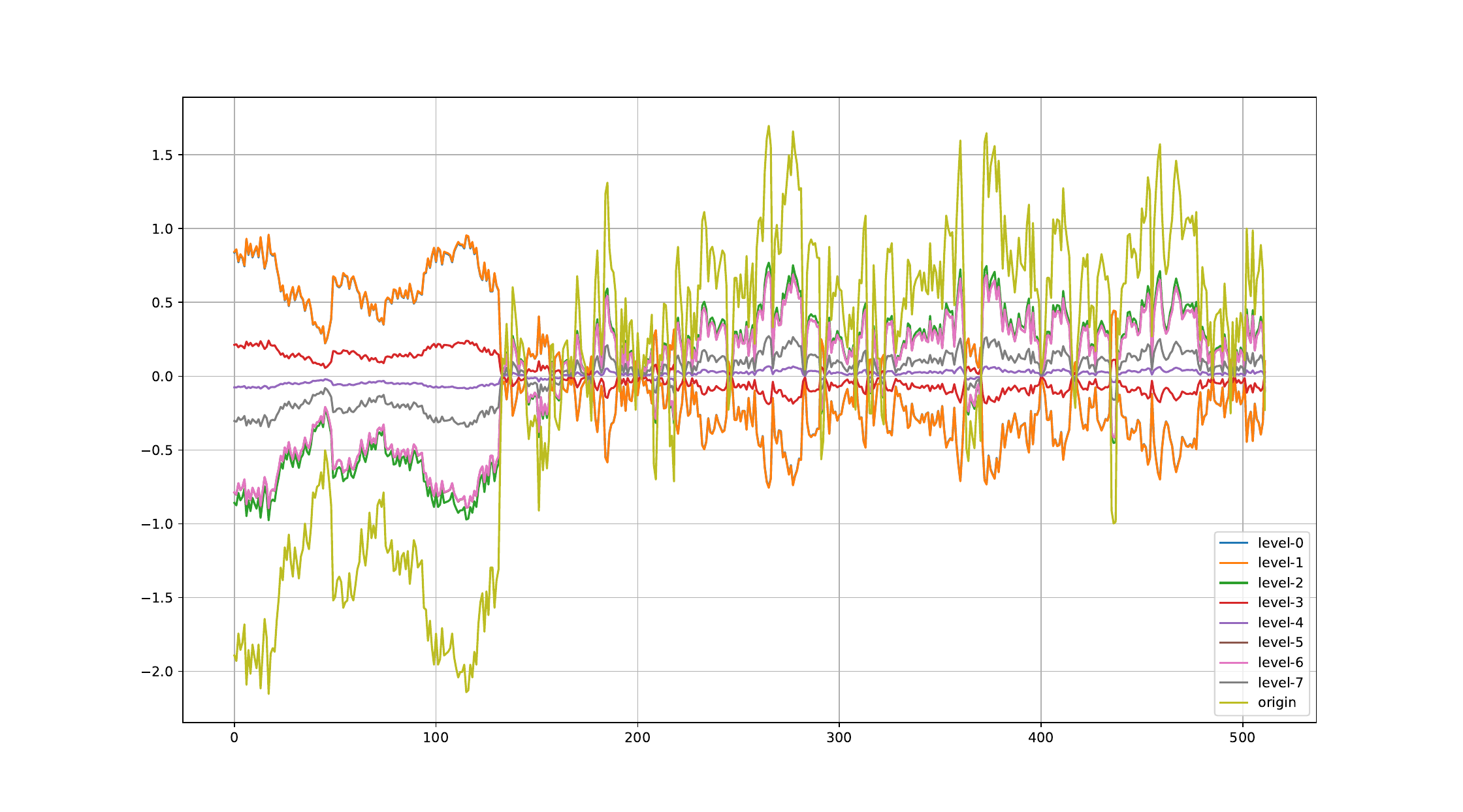}
  \caption{Multil-level property of ETTm2 series in the normalized time space by applying the learnd level-expansion matrix.}
  \label{fig:12}
\end{figure}
\begin{figure}[htbp]
\centering
  \includegraphics[width=\linewidth]{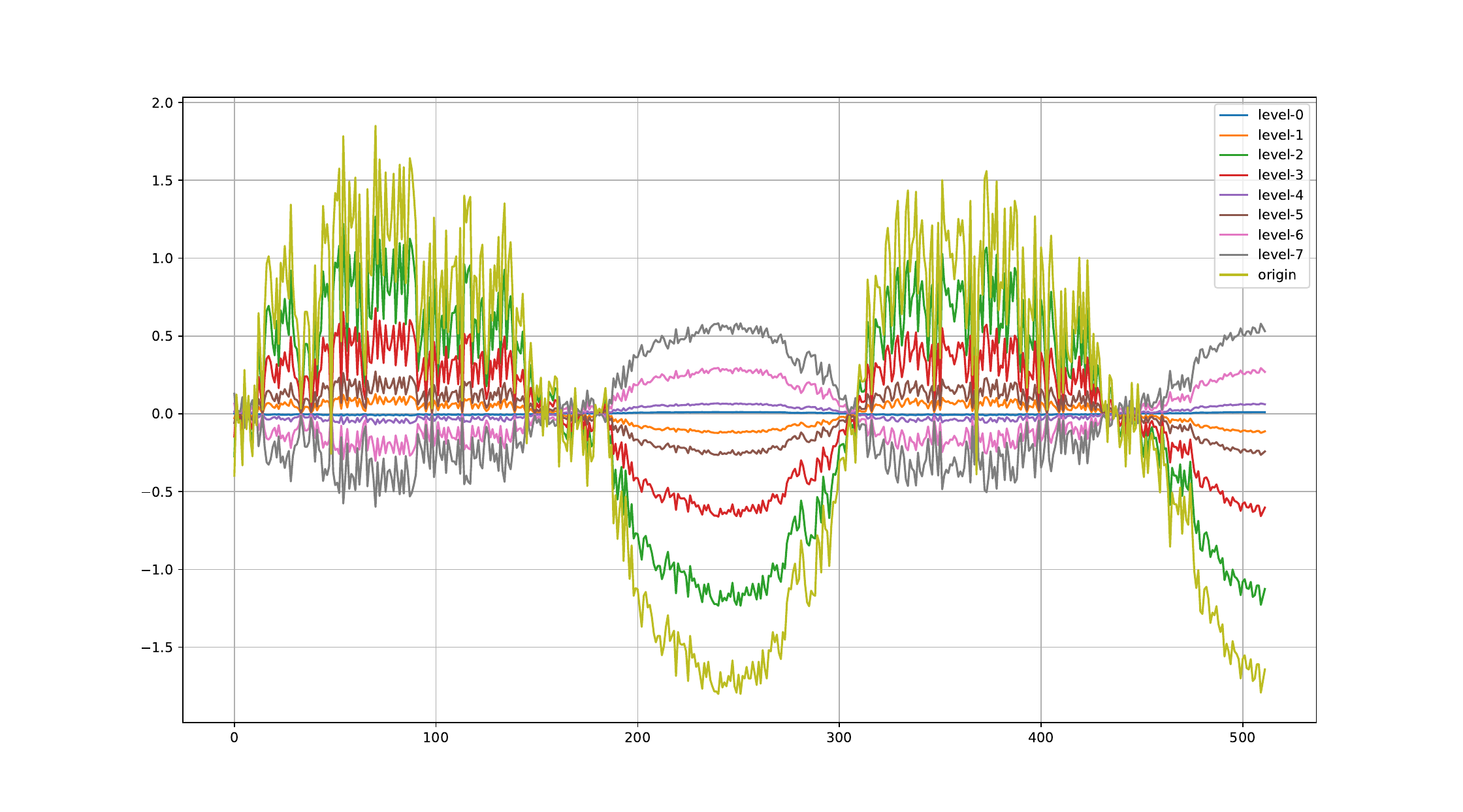}
  \caption{Multil-level property of PEMS04 series in the  normalized time space by applying the learnd level-expansion matrix.}
  \label{fig:13}
\end{figure}

This observation raises another question: is this multi-scale effect a result of the level expansion transform $\mathbf{M}_\text{L}$ ? To figure out it, we apply the transform again to $\mathcal{X}_0$, the original time series in the time domain, just after the instance normalization. We decompose it into the same number of sub-level series. As shown in Fig. \ref{fig:12} and Fig. \ref{fig:13}, these decomposed sub-series also exhibit different scales and frequencies. This demonstrates the effectiveness of the learned level expansion transform and suggests that we could leverage such transforms to generate new kinds of time series decompositions.
\begin{figure}[htbp]
\centering
  \includegraphics[width=\linewidth]{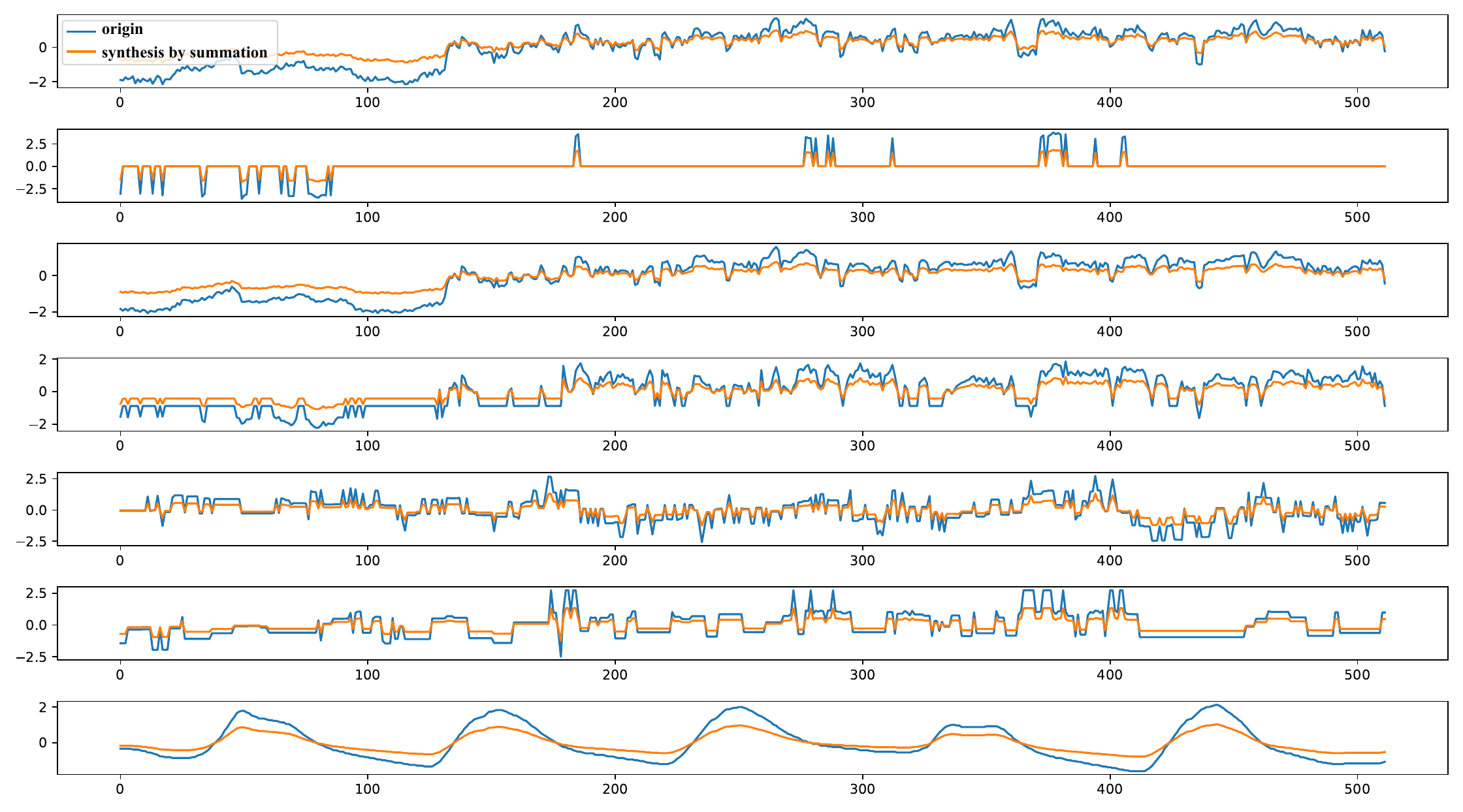}
  \caption{A summation of the sub-level series of the ETTm2 series in the normalized time space demonstrates that it approaches the original series, indicating that the linear decomposition has been achieved by the learned level-expansion matrix. Each subfigure represents a distinct variable of the ETTm2.}
  \label{fig:14}
\end{figure}
\begin{figure}[htbp]
\centering
  \includegraphics[width=\linewidth]{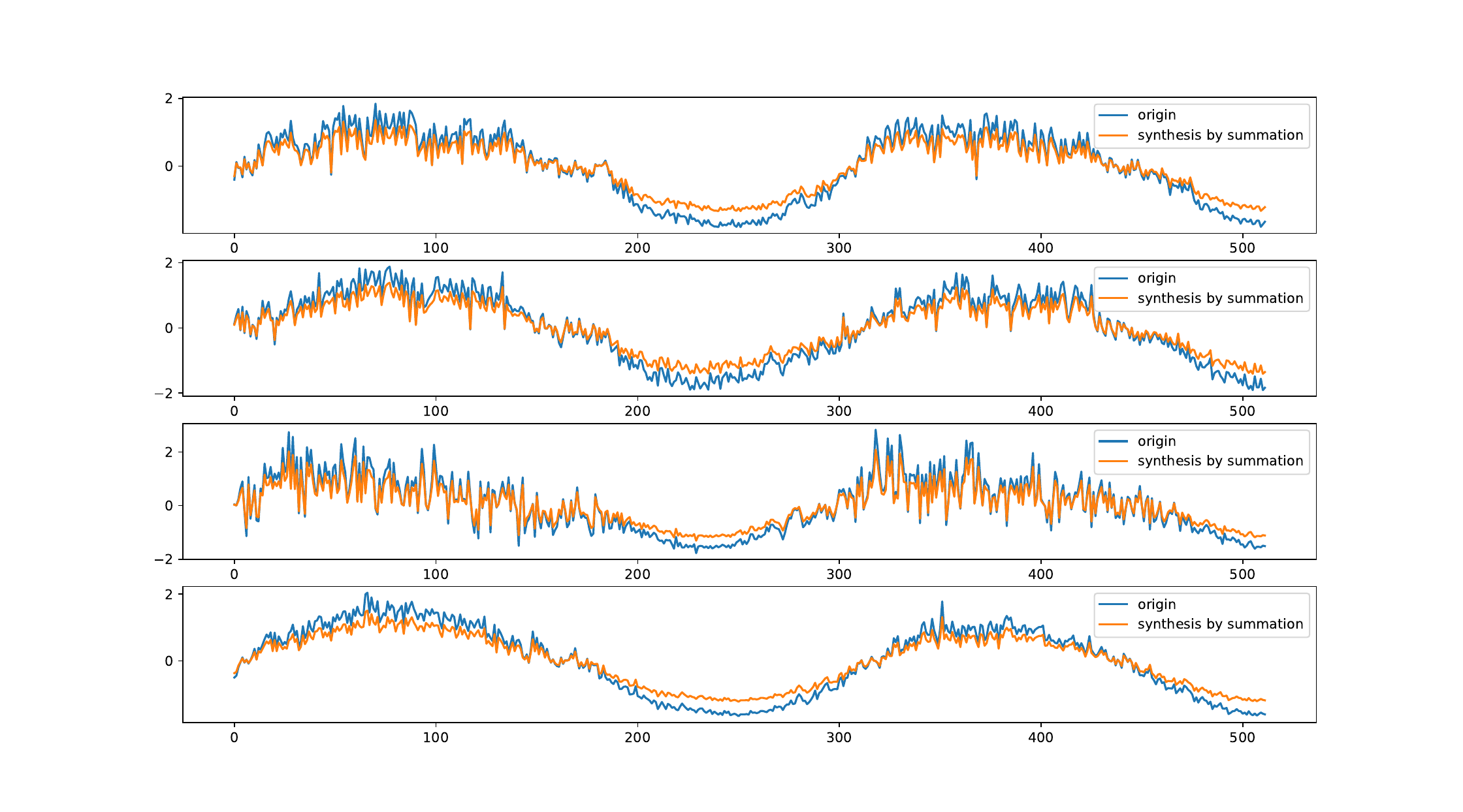}
  \caption{A summation of the sub-level series of the ETTm2 series in the normalized time space demonstrates that it approaches the original series, indicating that the linear decomposition has been achieved by the learned level-expansion matrix. The first four variables are demonstrated in the figure.}
  \label{fig:15}
\end{figure}

Finally, although we refer to the level expansion as a kind of time series decomposition, we need to check that these sub-level series are indeed the result of a certain decomposition. As a naive endeavor, we sum all the sub-level series of the decomposed $\mathcal{X}_0$ into a single series. Suprisingly, the summation result nearly reconstructs the original series, as shown in Fig. \ref{fig:14} and Fig. \ref{fig:15}. This validates that our model's $\mathbf{M}_\text{L}$ achieves a linear decomposition for time series. Moreover, it also suggests that these levels act as time bases, similar to the basis functions discussed in models like N-BEATs \cite{oreshkin2019nbits} and N-Hits \cite{challu2023nhits}. Broadly speaking, each sub-level captures a unique aspect of the series: high-frequency sub-series contribute to the coarse structure, while low-frequency sub-series refine finer details.  This may also help to explain why linear dependencies play a crucial role in time series modeling.
\section{Limitations and Future Work}
As discussed by current LTSF benchmark works \cite{qiu2024tfb}, no existing model can emerge as the best across all cases. While we are are pleased to find that TimeCapsule can consistently achieve SOTA performance on diverse datasets with competitive computation speeds, its flexibility also introduces many challenges such as the searching for the optimal hyperparameter. In the following, we elaborate on several limitations of our proposed model, which may stimulates further explorations and refinements.

\textbf{Compression Setting:} The model currently relies on fixed, hard-coded compression dimensions as hyperparameters, meaning that it is enforced to compress information into a predefined representation size. This constraint might restrict the model's full potential, especially for datasets that benefit from adaptive compression. Hence, exploring more mechanisms about time series decomposition and the compressibility of time series could lead to more flexible and effective architectures.

\textbf{Component Utilization:} TimeCapusle's modular structure sometimes reveals uneven utilization of its components. For instance, transformer blocks for compressed representation learning and dependency capture are highly effective in some cases but may go underused in others. Likewise, the dimension of linear projection within the MLP usually can significantly impact results. This variability points to a potential waste of computational resources when dealing with different datasets. Regarding to this issue, we will aim to analyze each component's role in the model and then distill the structure into a more compact form for practical applications.

\textbf{Generality:} The current version of TimeCapsule is specifically tailored to address one of the most important challenges in time series research—long-term time series forecasting (LTSF). Despite the shown initial classification results (see Appendix \ref{classifier}), its design and functionality are primarily centered around optimizing LTSF, which may inherently limit its expressiveness and effectiveness for other downstream tasks, such as classification and imputation. Exploring TimeCapsule's adaptability to these broader applications, including investigating structural adjustments to ensure scalability to build large pre-trained models and better accommodate diverse practical applications, represents an urgent modification direction for the proposed TimeCapsule.

\section{Investigation into the Potential for Classification}
\label{classifier}
In this section we take a first look at the potential of TimeCapsule in an alternative key time series application, classification, which has played a crucial role in many real world scenarios. In order to achieve this purpose with a minimal change to the model structure, we turn off the use of JEPA due to its unclear interpretation in classification.
\begin{table}[htbp]
\caption{Results for classification task. The classification accuracy $(\%)$ is recorded as the results below.} 
\label{tab:10}
\vskip 0.1in
\centering
\resizebox{.95\textwidth}{!}{\begin{tabular}{c|c|c|c|c|c|c|c|c|c}
\toprule
\diagbox{Datasets}{Methods} & Informer (\citeyear{zhou2021informer}) & Pyraformer (\citeyear{liu2021pyraformer}) & Autoformer (\citeyear{wu2021autoformer}) & FEDformer (\citeyear{zhou2022fedformer}) & iTransformer (\citeyear{liu2023itransformer}) & Dlinear (\citeyear{zeng2023dlinear}) & TiDE (\citeyear{tide}) & Timesnet (\citeyear{wu2022timesnet}) & TimeCapsule \\
\midrule
Heartbeat & 80.5 & 75.6 & 74.6 & 73.7 & 75.6 & 75.1 & 74.6 & 78.0 & 78.5 \\
FaceDetection & 67.0 & 65.7 & 68.4 & 66.0 & 66.3 & 68.0 & 65.3 & 68.6 & 70.2 \\
Handwriting & 32.8 & 29.4 & 36.7 & 28.0 & 24.2 & 27.0 & 23.2 & 32.1 & 27.0 \\
SelfRegulationSCP2 & 53.3 & 53.3 & 50.6 & 54.4 & 54.4 & 50.5 & 53.4 & 57.2 & 57.8 \\
EthanolConcentration & 31.6 & 30.8 & 31.6 & 28.1 & 28.1 & 32.6 & 27.1 & 35.7 & 32.0 \\
UWaveGestureLibrary & 85.6 & 83.4 & 85.9 & 85.3 & 85.9 & 82.1 & 84.9 & 85.3 & 88.8 \\
\midrule
Average Accuracy & 58.5 & 56.4 & 58.0 & 55.9 & 55.8 & 55.9 & 54.8 & 59.5 & 59.1 \\
\bottomrule
\end{tabular}}
\end{table}

We select six of the most challenging datasets used in \cite{wu2022timesnet}. The results in Table. \ref{tab:10} show that although all the modules designed in this paper are primarily dedicated to improving the generality and performance of the model in LTSF, it can still achieve competitive classification accuracies compared to other time series models in recent years, which further underscores the capability of TimeCapsule.
\newpage
\section{Visualizations of Forecasting}
\label{appen:vis}
\begin{figure}[h]
\subfigure[MLP (SAN) achieves $\text{MSE}=0.378$, this model explicitly predicts non-stationary statistics using SAN \cite{liu2024adaptivenorm}.]{
\begin{minipage}[hbtp]{\linewidth}
\includegraphics[width=\linewidth]{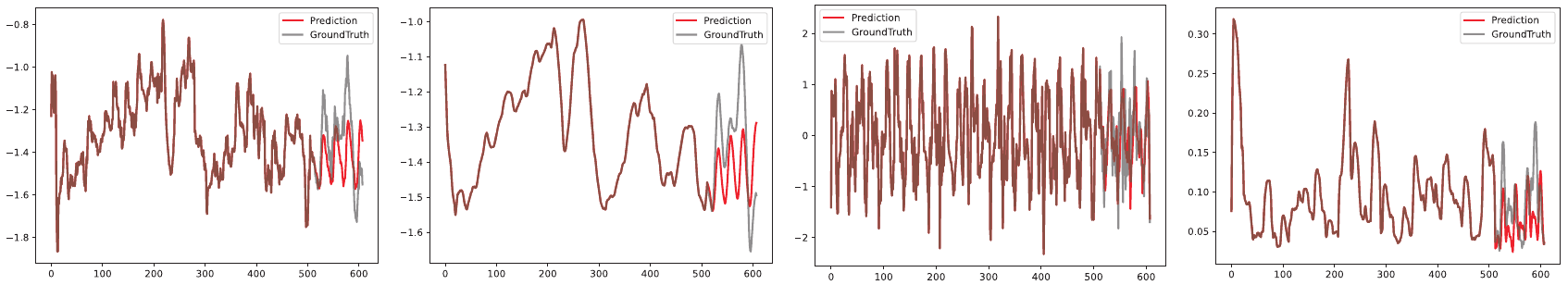}
\end{minipage}
}
\subfigure[iTransformer (version of non-stationary transformer) achieves $\text{MSE}=0.391$, this model explicitly models the non-stationary statistics within the non-stationary transformer \cite{liu2022nonsta}.]{
\begin{minipage}[hbtp]{\linewidth}
\includegraphics[width=\linewidth]{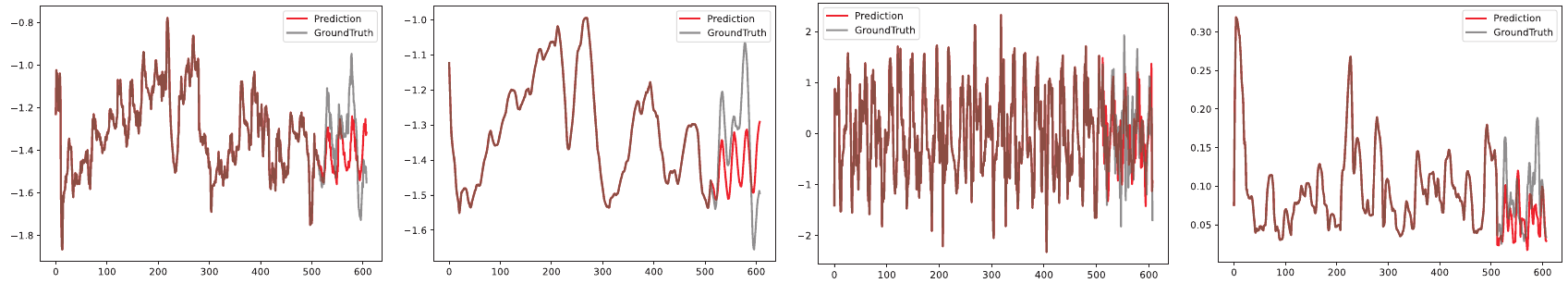}
\end{minipage}
}
\subfigure[MLP (RevIn) achieves $\text{MSE}=0.366$, this model is selected as a control group for the use of RevIn \cite{kim2021reversible}.]{
\begin{minipage}[hbtp]{\linewidth}
\includegraphics[width=\linewidth]{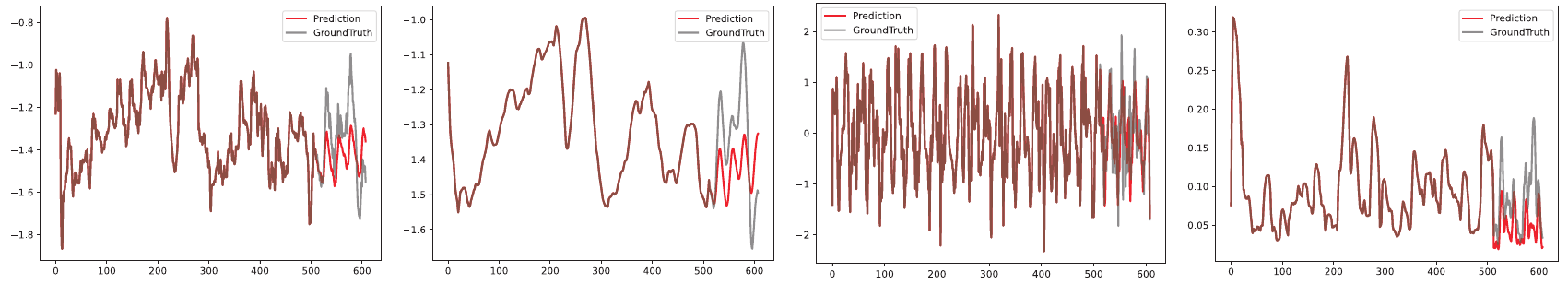}
\end{minipage}
}
\subfigure[TimeCapsule achieves $\text{MSE}=0.362$]{
\begin{minipage}[hbtp]{\linewidth}
\includegraphics[width=\linewidth]{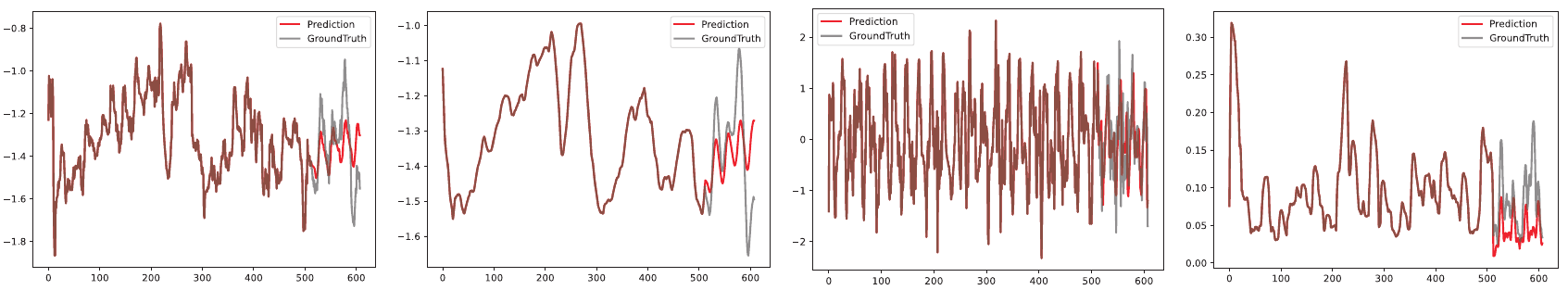}
\end{minipage}
}
\caption{We visualize ETTh1 dataset as a typical example to illustrate 512-96 predictions of different components (trend, stationary, deviation) of time series. All results are obtained under optimal settings of the source code.}
\label{fig:h}
\end{figure}

\begin{figure}[h]
\subfigure[MLP (SAN) achieves $\text{MSE}=0.378$, this model explicitly predicts non-stationary statistics using SAN \cite{liu2024adaptivenorm}.]{
\begin{minipage}[hbtp]{\linewidth}
\includegraphics[width=\linewidth]{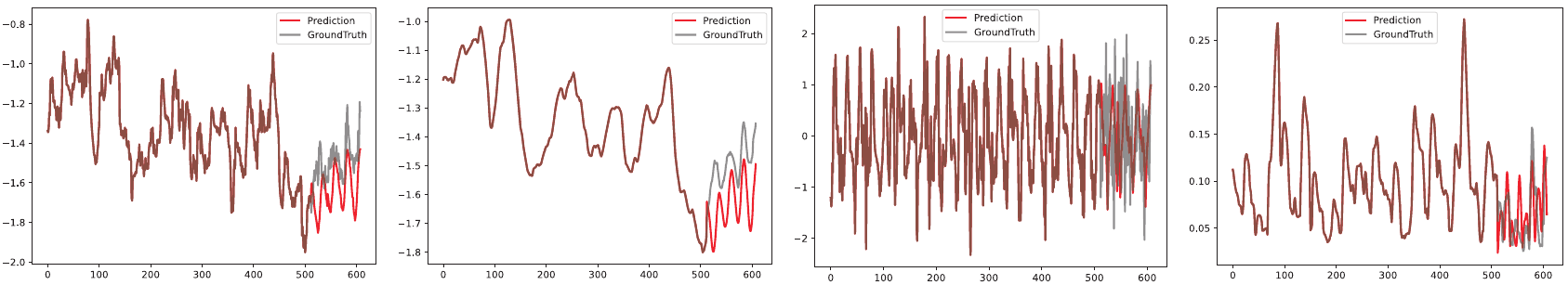}
\end{minipage}
}
\subfigure[iTransformer (version of non-stationary transformer) achieves $\text{MSE}=0.391$, this model explicitly models the non-stationary statistics within the non-stationary transformer \cite{liu2022nonsta}.]{
\begin{minipage}[hbtp]{\linewidth}
\includegraphics[width=\linewidth]{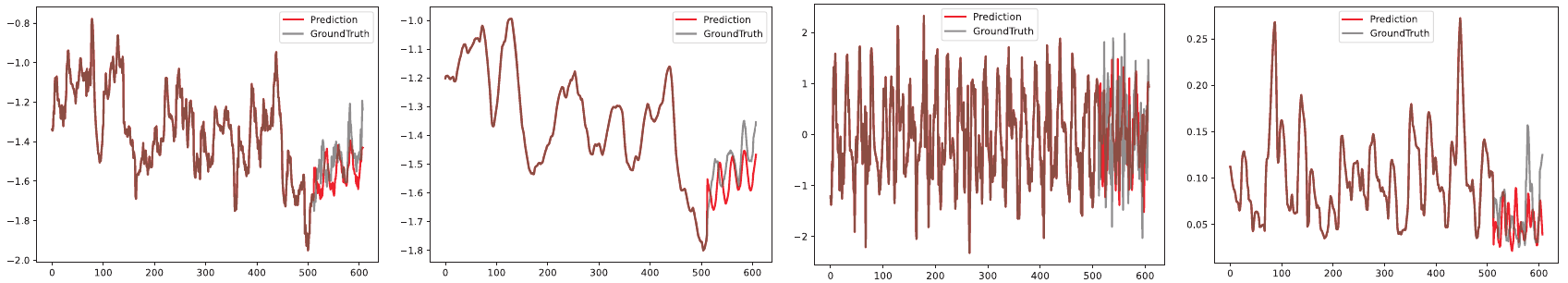}
\end{minipage}
}
\subfigure[MLP (RevIn) achieves $\text{MSE}=0.366$, this model is selected as a control group for the use of RevIn \cite{kim2021reversible}.]{
\begin{minipage}[hbtp]{\linewidth}
\includegraphics[width=\linewidth]{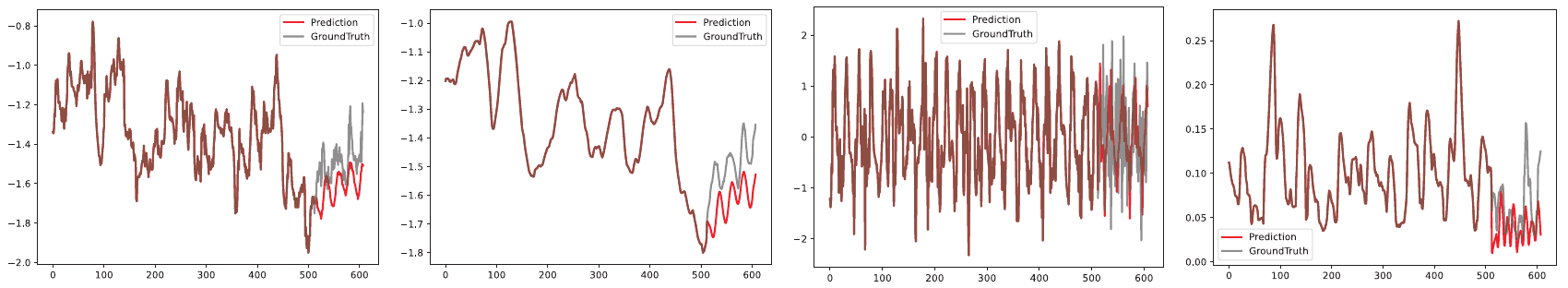}
\end{minipage}
}
\subfigure[TimeCapsule achieves $\text{MSE}=0.362$]{
\begin{minipage}[hbtp]{\linewidth}
\includegraphics[width=\linewidth]{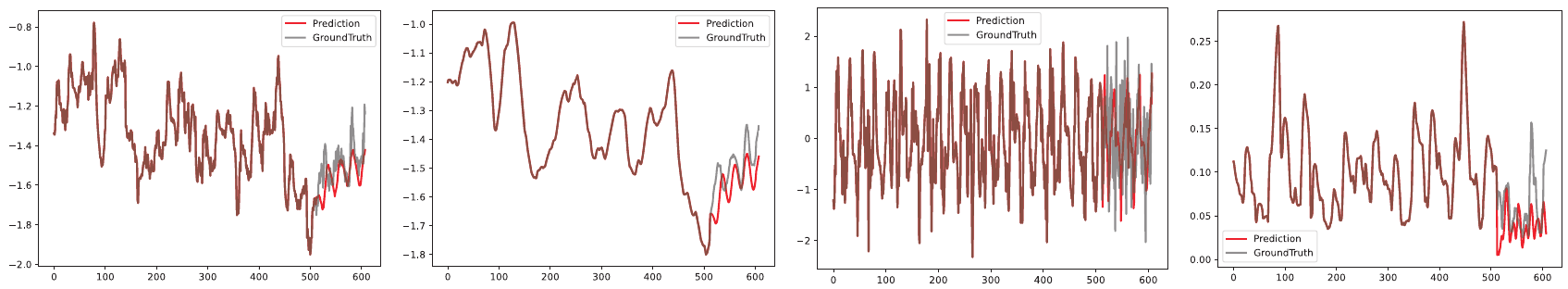}
\end{minipage}
}
\caption{We visualize ETTh1 dataset as a typical example to illustrate 512-96 predictions of different components (trend, stationary, deviation) of time series. All results are obtained under optimal settings of the source code.}
\label{fig:z}
\end{figure}

\begin{figure}[h]
\centering
  \includegraphics[scale=0.4]{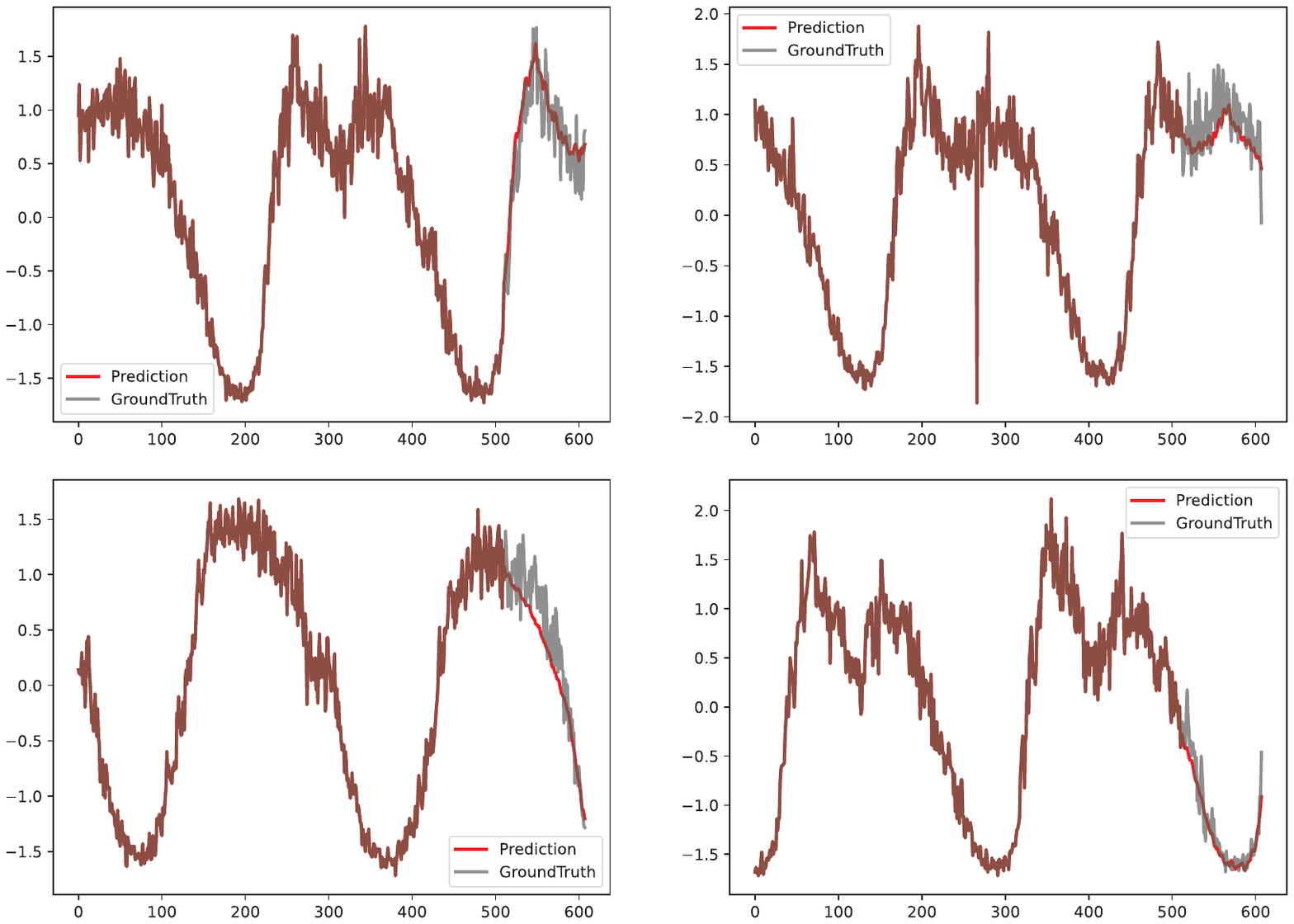}
  \caption{Prediction on the PEMS04 dataset, with lookback window 512 and forecast length 96.}
\end{figure}
\begin{figure}[h]
\centering
  \includegraphics[scale=0.4]{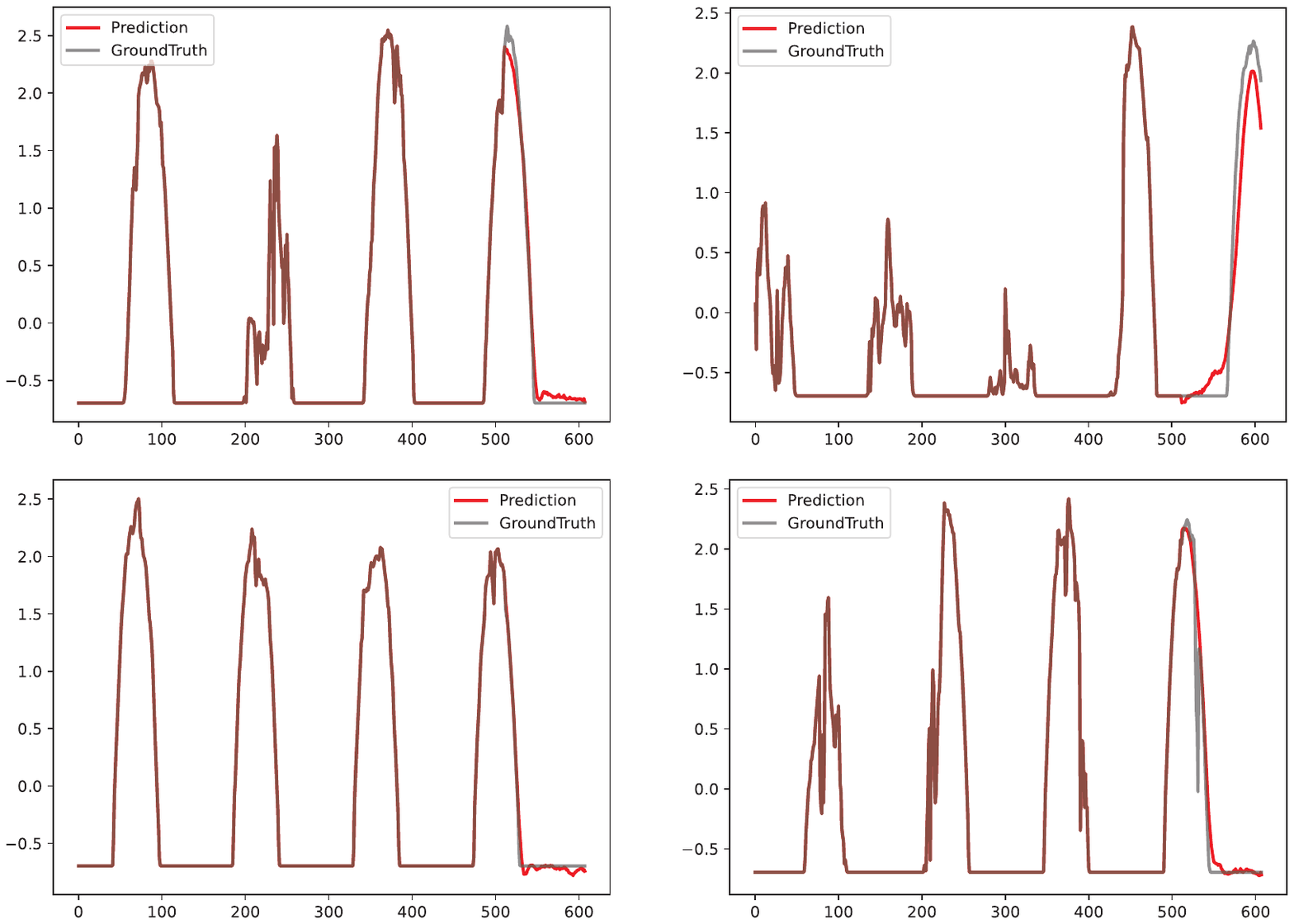}
  \caption{Prediction on the Solar dataset, with lookback window 512 and forecast length 96.}
\end{figure}
\begin{figure}[h]
\centering
  \includegraphics[scale=0.4]{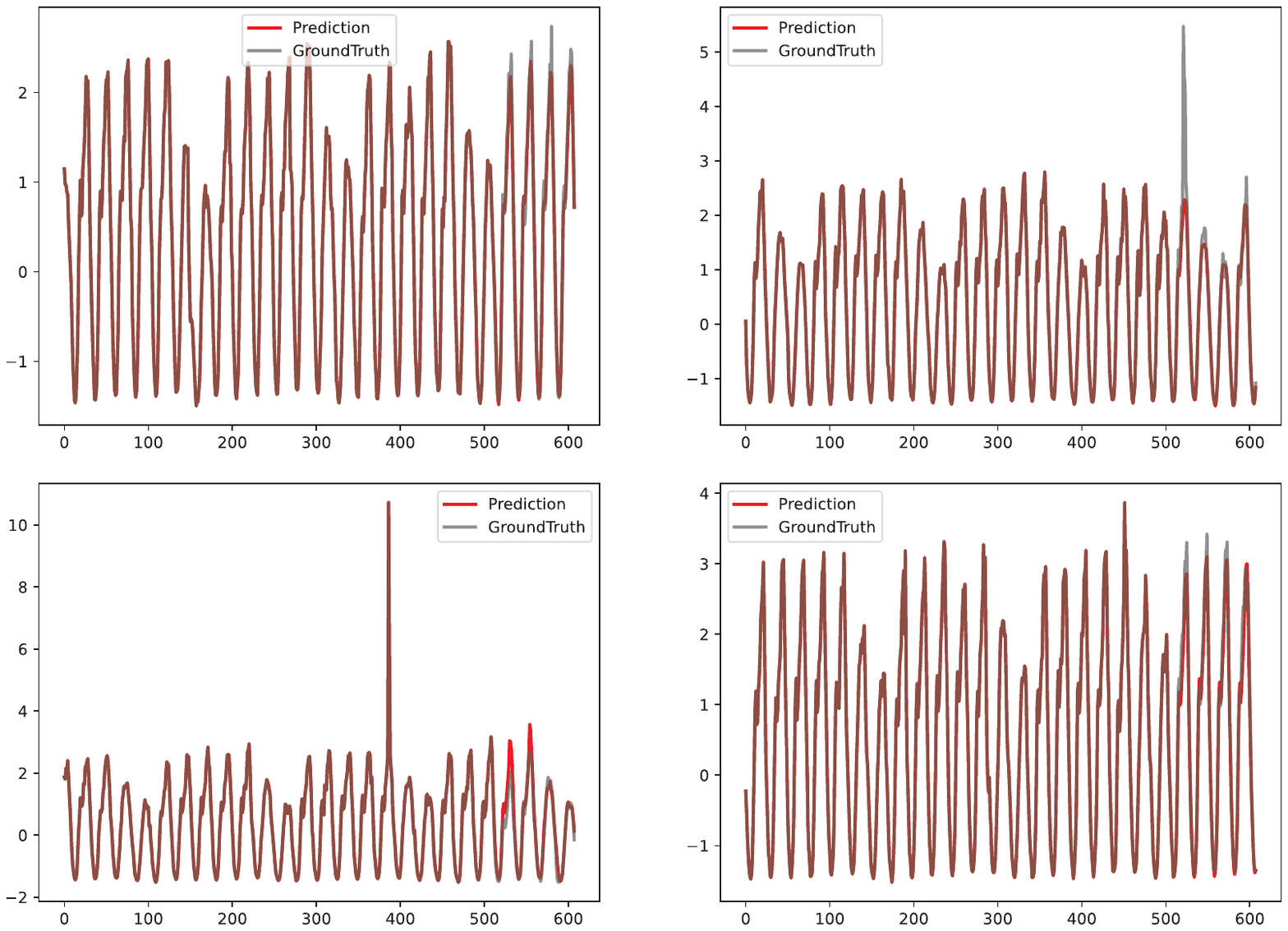}
  \caption{Prediction on the Traffic dataset, with lookback window 512 and forecast length 96.}
\end{figure}
\begin{figure}[h]
\centering
  \includegraphics[scale=0.4]{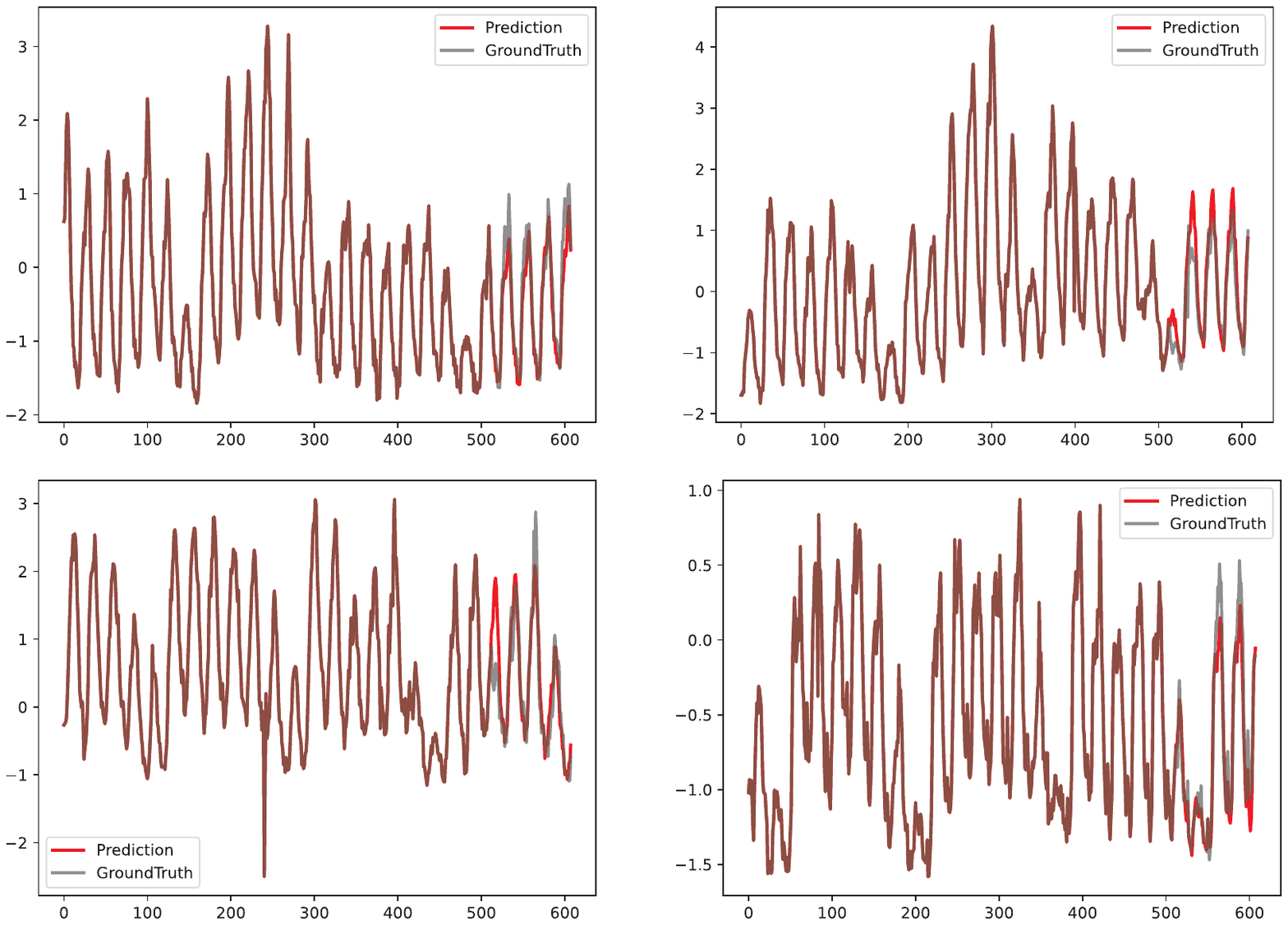}
  \caption{Prediction on the Electricity dataset, with lookback window 512 and forecast length 96.}
\end{figure}
\end{document}